\documentclass[conference]{IEEEtran}
\IEEEoverridecommandlockouts
\usepackage{amsmath,amssymb,amsfonts}
\usepackage{algorithmic}
\usepackage{graphicx}
\usepackage{textcomp}
\usepackage{xcolor}
\usepackage{optidef}
\usepackage{lipsum}
\usepackage{multirow}
\usepackage{readarray}
\usepackage{rotating}
\usepackage[square,sort,comma,numbers,compress]{natbib}
\usepackage{microtype}
\usepackage[inline]{enumitem}
\newcommand{\citefull}[1]{\citeauthor{#1}~\cite{#1}}

\usepackage[commentmarkup=todo,todonotes={textsize=tiny,textwidth=1.3cm}]{changes}
\definechangesauthor[name=Martin, color=green]{MK}
\definechangesauthor[name=Peter, color=blue]{PZ}
\usepackage{amsmath}
\usepackage{pgfplots}
\usepackage{pgfplotstable}
\pgfplotsset{compat=1.15}
\usepackage{placeins}
\usepackage{pdflscape}
\usepackage{adjustbox}
\usepackage{tkz-fct}
\usetikzlibrary{arrows}
\usetikzlibrary{circuits}
\usetikzlibrary{decorations}
\usepackage{hhline,multirow}
\usepackage{rotating}
\usepackage{booktabs}
\usepackage{comment}
\usepackage{subcaption}
\usepackage{glossaries-prefix}
\usepackage{framed}
\usepackage[ruled,vlined]{algorithm2e}
\usepackage[bookmarks=false]{hyperref}
\usepackage[noabbrev,capitalize]{cleveref}
\usepackage{float}
\newcolumntype{C}[1]{>{\centering\arraybackslash}p{#1}} % centering column type with fixed width
\newcolumntype{R}[1]{>{\raggedleft\arraybackslash}p{#1}} % right aligned column type with fixed width
\newcolumntype{L}[1]{>{\raggedright\arraybackslash}p{#1}} % left aligned column type with fixed width

\tikzset{
	png export/.style={
		external/system call/.add=
		{}
		{; convert -density 300 -transparent white "\image.pdf" -quality 100 "\image.png";},
		/pgf/images/external info,
		/pgf/images/include external/.code={%
			\includegraphics
			[width=\pgfexternalwidth,height=\pgfexternalheight]
			{##1.png}%
		}
	}
}
\tikzset{
	eps export/.style={
		external/system call/.add=
		{}
		{; convert -density 300 "\image.pdf" -quality 100 "\image.eps";},
		/pgf/images/external info,
		/pgf/images/include external/.code={%
			\includegraphics
			[width=\pgfexternalwidth,height=\pgfexternalheight]
			{##1.eps}%
		}
	}
}
\IEEEoverridecommandlockouts

\def\BibTeX{{\rm B\kern-.05em{\sc i\kern-.025em b}\kern-.08em
    T\kern-.1667em\lower.7ex\hbox{E}\kern-.125emX}}
\begin{document}

\title{FAT: Training Neural Networks for Reliable Inference Under Hardware Faults }
%\IEEEauthorrefmark{1}
\author
{
\IEEEauthorblockN{Ussama Zahid\IEEEauthorrefmark{1}, Giulio Gambardella\IEEEauthorrefmark{1}, Nicholas J. Fraser\IEEEauthorrefmark{1}, Michaela Blott\IEEEauthorrefmark{1}, Kees Vissers\IEEEauthorrefmark{2}}
\IEEEauthorblockA{\IEEEauthorrefmark{1}Xilinx Research Labs, Dublin, Ireland, Email: \{ussamaz, giuliog, nfraser, mblott\}@xilinx.com}
\IEEEauthorblockA{\IEEEauthorrefmark{2}Xilinx Research Labs, San Jose, USA, Email: keesv@xilinx.com}
}

\maketitle

% Capital letters are for proper nouns ONLY!

% Machine learning-related:
\newacronym{ANN}{ANN}{artificial neural network}
\newacronym{DNN}{DNN}{deep neural network}
\newacronym{QNN}{QNN}{quantized neural network}
\newacronym{FAT}{FAT}{fault-aware training}
\newacronym{SAT}{SAT}{standard training}
\newacronym{CNN}{CNN}{convolutional neural network}
\newacronym{OFM}{OFM}{output feature map}
\newacronym{MLP}{MLP}{multi-layer perceptron}
\newacronym{FC}{FC}{fully connected}
\newacronym{NN}{NN}{neural network} % Do we need this and ANN? I'd prefer you to pick on and to use throughout

% Hardware-related
\newacronym{FPGA}{FPGA}{field programmable gate arrays}
\newacronym{MAC}{MAC}{multiply-accumulate}
\newacronym{PE}{PE}{processing element}
\newacronym{MVTU}{MVTU}{matrix vector threshold unit}

% Safety related
\newacronym{RAMS}{RAMS}{reliability, availability,  maintainability and safety}
\newacronym{FMEDA}{FMEDA}{failure, modes, effects, and diagnostic analysis}
\newacronym{SEE}{SEE}{single-event-effect}
\newacronym{TMR}{TMR}{triple modular redundancy}
\newacronym{HCI}{HCI}{hot-carrier injection}
\newacronym{NBTI}{NBTI}{negative bias temperature instability}

% Miscellaneous
\newacronym{FPS}{FPS}{frames per second}

\begin{abstract}
\Glspl{DNN} are state-of-the-art algorithms for multiple applications, spanning from image classification to speech recognition.
While providing excellent accuracy, they often have enormous compute and memory requirements.
As a result of this, \glspl{QNN} are increasingly being adopted and deployed especially on embedded devices, thanks to their high accuracy,
but also since they have significantly lower compute and memory requirements compared to their floating point equivalents.
\Gls{QNN} deployment is also being evaluated for safety-critical applications, such as automotive, avionics, medical or industrial.
These systems require functional safety, guaranteeing failure-free behaviour even in the presence of hardware faults.
In general fault tolerance can be achieved by adding redundancy to the system, which further exacerbates the overall computational demands and makes it difficult to meet  the power and performance requirements. 
%Exploitation of \glspl{DNN} inherent feature is vital to decrease cost of functional safety extensions in acceleration hardware. 
In order to decrease the hardware cost for achieving functional safety, it is vital to explore domain-specific solutions which can exploit the inherent features of \glspl{DNN}.
In this work we present a novel methodology called \gls{FAT}, which includes error modeling during \gls{NN} training, to make \glspl{QNN} resilient to specific fault models on the device.
Our experiments show that by injecting faults in the convolutional layers during training, highly accurate \glspl{CNN} can be trained which exhibits much better error tolerance compared to the original.
Furthermore, we show that redundant systems which are built from \glspl{QNN} trained with \gls{FAT} achieve higher worse-case accuracy at lower hardware cost.
This has been validated
%We validated our approach 
for numerous classification tasks including CIFAR10, GTSRB, SVHN and ImageNet.
\end{abstract}


\section{Introduction}
\label{sec:introduction}

\glspl{CNN} have recently gained significantly in popularity due to their high accuracy in numerous applications such as image classification~\cite{Krizhevsky:2012:ICD:2999134.2999257}, object detection and localization~\cite{redmon2015look}, image  segmentation~\cite{badrinarayanan2015segnet,ronneberger2015unet}. In some applications, such as image classification, \glspl{CNN} can even outperform human beings~\cite{he2015delving}.
Nonetheless, the push for accuracy increase came at the cost of escalating  computational and memory demands.
This in turn, started many branches of research targeting reducing the deployment cost of \glspl{CNN}.
%Recently there has been an immense effort in reducing the computational cost and memory requirements of \glspl{CNN} in order to deploy them in different applications i.e. smartphones, \glspl{FPGA}, etc.
These techniques include: direct quantization~\cite{gysel2016ristretto}, quantization-aware training~\cite{binary_net, rastegari2016xnornet, zhou2016dorefanet, cai2017deep}, pruning and compression~\cite{han2015deep}, using less intensive layers (depth-wise separable convolutions~\cite{MobileNet}), non-arithmetic layers (e.g., ShiftNet~\cite{Shiftnet}).
Reduced precision \glspl{DNN} reduce the memory footprint compared to a full precision one, sometimes incurring in modest drop in accuracy, which can be compensated with network topology changes~\cite{scalingBNN}.
Reduced precision \glspl{DNN} have weights and activations for which possible values are restricted within a few bits, thus requiring smaller adders and multipliers in hardware for the involved \gls{MAC} operations as compared to floating point \glspl{DNN}.
This is especially true on reconfigurable hardware, on which it is possible to tailor hardware to the needed precision, enabling scaling of parallelism when hardware cost is decreased. 
Additionally, quantization makes it easier to deploy \glspl{DNN} on small embedded devices for real-time applications~\cite{DBLP:journals/corr/TripathiDKBN17}.
FINN~\cite{finn} is one of the frameworks which demonstrates an efficient way of deploying \glspl{QNN} on \gls{FPGA} which leverages quantization to achieve high throughput in terms of \gls{FPS}.

%In order to use an electronic device in a safety critical application, its dependability must be evaluated, usually composed of \gls{RAMS}.
In  order  to  use  an  electronic  device  in  a  safety  critical application, its dependability must be evaluated. This consists of four components: \gls{RAMS}~\cite{RAMS}.
The safety analysis and its evaluation is regulated by the standards depending on the application domain (e.g., IEC-61508 for industrial systems, ISO-26262 for road vehicles and EN 50126/8/9 for rail transport), with safety levels proportional to the criticality of the specific task.
\Gls{FMEDA}~\cite{FMEDA} have to be evaluated for each of the components of the design in order to model the safety features of the system.

During deployment, when a \gls{DNN} is running on some execution hardware, there is a non-zero probability of the hardware malfunctioning.
This malfunctioning can be caused by permanent or transient hardware faults.
Common sources of transient errors are direct or indirect ionization, mostly by means of heavy ions.
This is especially true for avionics or space applications where protons are likely to hit hardware circuit and generate transient effects.
Permanent faults are instead commonly linked with electrical failures in digital circuits, like electromigration, aging (due to \gls{HCI} or \gls{NBTI}).
Electrical faults will often cause a disconnection in wiring along with a misbehaviour in the functioning of a transistor, which can lead to open or short circuits in the output of a logic gate.
Two common errors which can arise in the computation from hardware faults are \glspl{SEE} (transient) and stuck-at (permanent).
The result of \glspl{SEE} is usually a single-bit error in a stream of data, or a bit-flip in a memory array.
Stuck-at errors usually correspond to some larger fault within the hardware, which causes the output of a particular circuit to have a singular value.
A stuck@x fault refers to this type of fault, where a given value in a circuit is always stuck at the value x, rather than the value intended by the circuit itself.
Note that on \glspl{FPGA}, if no measures like scrubbing are adopted, permanent errors can occur from transient fault effecting the \gls{FPGA} configuration memory. 
%Depending on the amount parallelism of the hardware and how a particular layer is mapped to that hardware, decides how a particular fault can present itself as different errors in the output of the given layer.
In order to safely adopt \glspl{DNN} in safety critical applications, it is required that they must be tolerant to different fault models.
\Glspl{QNN} tolerance measurements can be done through \textit{error injection}, which can be used to evaluate the accuracy in the presence of possible errors, as presented by~\citefull{FT_QNN} and~\citefull{SEU_BNN}.
In this work, we propose a novel training method which leverages error modelling during the training process of \glspl{QNN} in order to make them resilient to particular errors when deployed for inference in the field. 
The main contributions of this paper are:
\begin{itemize}
    \item an efficient methodology (\gls{FAT}) to train error tolerant \glspl{QNN}; 
    \item an evaluation of our methodology compared to Dropout2D~\cite{tompson2015efficient} as an alternative method to introduce fault tolerance in the \glspl{QNN}; and
    \item a hardware cost vs. worst case error analysis of \gls{FAT} trained \glspl{QNN} vs. \glspl{QNN} trained with standard quantized neural network training techniques, which we denote as \gls{SAT}.
\end{itemize}
This paper is organised as follows:
%\begin{itemize}
%    \item \cref{sec:background} describes different types of \gls{QNN} accelerators and introduces some error models with can arise from faults in \gls{QNN} hardware;
%    \item \cref{sec:related work} discusses prior work related to the fault tolerance of \glspl{QNN} and introduces a motivating example for this work;
%    \item \cref{sec:FAT} introduces our proposed methodology; and finally
%    \item \cref{sec:experimental,sec:observations,sec:conclusion} contain results, discussions and conclusions respectively.
%\end{itemize}
\cref{sec:background} describes different types of \gls{QNN} accelerators and introduces some error models with can arise from faults in \gls{QNN} hardware;
\cref{sec:related work} discusses prior work related to the fault tolerance of \glspl{QNN} and introduces a motivating example for this work;
\cref{sec:FAT} introduces our proposed methodology; and finally
\cref{sec:experimental,sec:observations,sec:conclusion} contain results, discussions and conclusions respectively.

\section{Background}
\label{sec:background}

\Glspl{CNN} are usually composed of a sequence of layers, each with its own characteristics such as: feature map sizes, channels and filters.
We refer the readers to the works by~\citefull{Mittal2018} and~\citefull{vkb18} for an exhaustive list of \gls{CNN} accelerators targeting programmable logic.
In this paper, we adopt the \gls{QNN} accelerators code-named FINN~\cite{finn, finn-r} which are publicly available on GitHub~\cite{BNN-PYNQ-REPO}.
\citefull{finn-r} presented two different hardware architectures, namely \textit{dataflow} and \textit{loopback}.
In \textit{dataflow}, all layers are concurrently offloaded in programmable logic, with each layer mapped to its own array of \glspl{PE}.
Each array executes a single layer in a folded fashion, meaning each \gls{PE} is computing 1 or more outputs (e.g., an entire channel of an \gls{OFM}) of a \textit{single} layer. 

In \textit{loopback}, a single array of \glspl{PE} computes the \gls{CNN} output, computing each layer in sequential manner.
This implies that each PE is computing 1 or more output of \textit{multiple} layers.

Independently from the hardware architecture, possible faults in any of the \glspl{PE} can present themselves as errors in the intermediate values of the \gls{CNN} during inference time.
These errors are likely to have different properties based on the design of the \gls{CNN} execution engine and the nature of the fault which occurs.
%Fault in any hardware blocks on the architectures can be reflected as an error model in the \gls{CNN}.

The portion of the network affected by a corrupted \gls{PE} depends upon the scheduling of the \gls{CNN} and the architecture of the hardware accelerator.  
For example in FINN \textit{dataflow}, each \gls{PE} is scheduled for computing one or more entire \gls{OFM} channels of only one layer. 
If a \gls{PE} of a specific layer is faulty, the \gls{OFM} channels (of that particular layer) which are being calculated by the faulty \gls{PE}, are potentially erroneous.
Whereas in the \textit{loopback} architecture, \glspl{PE} are being used to execute multiple layers. 
Hence a fault in any of the \glspl{PE} will effect one or more \gls{OFM} channel(s) of \textit{multiple} layers. 
Similarly, if with a different scheduling each \gls{PE} is computing a particular pixel for all channels, faulty \gls{PE} would result in same pixel being erroneous for all \gls{OFM} channels. 
In general, depending on the hardware architecture and on the scheduling, different  models should be used to better represent possible permanent errors of the \gls{CNN} accelerator. 
%In case of \glspl{SEE} in dataflow, a single pixel value would be at error. 

%In the remainder of the paper we focus on single channel stuck@ errors and same pixel in all channels stuck@ as shown in \cref{fig:faulty PE}. The errors which cause the entire output channel of a particular layer or all channels in the same pixel contain the same value.
%Channel stuck@ and same pixel stuck@ respectively, models the faults in architectures where each \gls{PE} calculates one or more complete \gls{OFM} channels and all channels of same pixel. 
In the remainder of the paper we will focus on two main error models: single channel stuck@ errors and same pixel in all channels stuck@. The former cause an entire output channel of a particular layer being stuck-at, while the latter has all channels in the same pixel containing the same value.
%Channel stuck@ and same pixel stuck@ respectively, models the faults in architectures where each \gls{PE} calculates one or more complete \gls{OFM} channels and all channels of same pixel. 

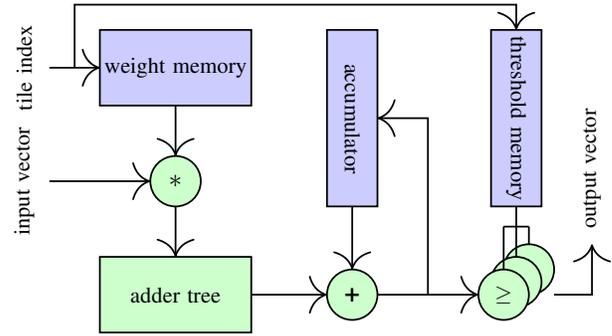
\begin{figure}
\centering
\resizebox{\columnwidth}{!}{
\begin{tikzpicture}

    \filldraw[fill=green!20, draw=black, line width = 0.07mm] (0,0) rectangle (0.6, 0.3) node [midway, scale = 0.25] {adder tree};
    % product to product
    \draw[->, color=black, line width=0.07mm] (0.3, 0.5) -- (0.3, 0.3);    
    % product to sum
    \draw[->, color=black, line width=0.07mm] (0.6, 0.15) -- (0.9, 0.15);    
    
    \filldraw[fill=green!20, draw=black, line width=0.07mm](0.3,0.6) circle (0.1);
    \draw[color=black](0.3,0.6) node [scale = 0.3]{\textbf{$*$}};
    \draw(-0.3,0.6) node[scale=0.25] {\begin{turn}{90}input vector\end{turn}};
    % weight to product
    \draw[->, color=black, line width=0.07mm] (0.3, 0.9) -- (0.3, 0.7);
    % input vector
    \draw[->, color=black, line width=0.07mm] (-0.2, 0.6) -- (0.2, 0.6);
    
   	\filldraw[fill=blue!20, draw=black, line width = 0.07mm] (0,0.9) rectangle (0.6, 1.2) node [midway, scale = 0.25] {weight memory};
    \draw(-0.3,1.05) node[scale=0.25] {\begin{turn}{90}tile index\end{turn}};
    % index to weight
    \draw[->, color=black, line width=0.07mm] (-0.2, 1.05) -- (0, 1.05);
    % from index to threshold memory
    \draw[->, color=black, line width=0.07mm] (-0.1, 1.05) -- (-0.1, 1.3) -- (1.65,1.3) -- (1.65,1.2);

    \filldraw[fill=green!20, draw=black, line width=0.07mm](1,0.15) circle (0.1);
    \draw[color=black](1,0.15) node [scale = 0.3]{\textbf{+}};
    % accumulator to sum
    \draw[->, color=black, line width=0.07mm] (1, 0.5) -- (1, 0.25);
    % sum to threshold comparison
    \draw[->, color=black, line width=0.07mm] (1.1, 0.15) -- (1.5, 0.15);
    % sum to accumulator
    \draw[->, color=black, line width=0.07mm] (1.3, 0.15) -- (1.3, 0.85) -- (1.1, 0.85);

   	\filldraw[fill=blue!20, draw=black, line width = 0.07mm] (0.9, 0.5) rectangle (1.1, 1.2) node [midway, scale=0.25]{\begin{turn}{-90}accumulator\end{turn}};	

	\filldraw[fill=blue!20, draw=black, line width = 0.07mm] (1.55, 0.5) rectangle (1.75, 1.2) node [midway, scale=0.25]{\begin{turn}{-90}threshold memory\end{turn}};
	
	\foreach \a in {2,...,0}
		\def \x { \a * 0.05}	
        \filldraw[fill=green!20, draw=black, line width=0.07mm](1.6+\x,0.15+\x) circle (0.1);

    \draw[color=black](1.6,0.15) node [scale = 0.25]{$\geq$};
    % threshold to comparison
    \draw[-, color=black, line width=0.07mm] (1.65, 0.5) -- (1.65, 0.3);
    \draw[-, color=black, line width=0.07mm] (1.6, 0.42) -- (1.7, 0.42);

    \draw[-, color=black, line width=0.07mm] (1.6, 0.25) -- (1.6, 0.42);
    \draw[-, color=black, line width=0.07mm] (1.7, 0.35) -- (1.7, 0.42);

    % comparison to output
    \draw[->, color=black, line width=0.07mm] (1.8, 0.15) -- (1.95, 0.15) -- (1.95, 0.35);
    \draw(1.95,0.65) node[scale=0.25] {\begin{turn}{90}output vector\end{turn}};

\end{tikzpicture}
}
\caption{Matrix Vector Threshold Unit (MVTU) block diagram, core component of a PE in FINN} \label{fig:mvtu}
\vspace{-0.5cm}
\end{figure}

%Channel stuck@ models the faults in an architectures where each \gls{PE} calculates one or more complete \gls{OFM}, while pixel stuck@ models the faults in a \gls{PE} which computes all channels of the same pixel. 
With the FINN architecture, channel stuck@ models the possible faults within a single \gls{PE}, composed of MAC and activation block (called \gls{MVTU}) as shown in \cref{fig:mvtu}. 
In the \gls{MVTU} architecture, a faulty bit in the input vector, adder tree, weight memory or in the multiplier will most likely affect the accumulator value. Faults in the threshold memory or in the comparators would have effects on the result of the output vector. 
All of the above faults, if not masked, could thus result in an error in the output activation. In the worst case scenario, the error would be propagated to the entire output channel, given the adopted scheduling. 
On the other hand, pixels stuck@ in FINN models the faults in the communication between layers, since each layer produces streams of all channels for a single pixel, and pixels are generated in a raster order. 
A possible fault in the communication FIFO would then result in a complete pixel being compromised.

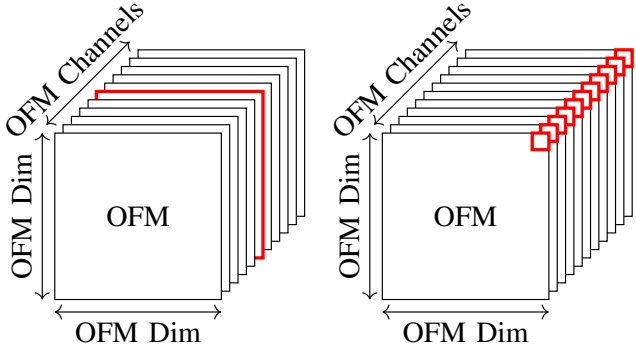
\begin{figure}
%\centering
\resizebox{\columnwidth}{!}
{
\begin{tikzpicture}
	%modules
	\foreach \a in {10,...,6}
		\def \x { \a / 10}
		\filldraw[fill=white, draw=black] (0 + \x,0 + \x) rectangle (2 + \x,2 + \x);
	
	\filldraw[fill=white, draw=red, line width = 0.4mm] (0.5, 0.5) rectangle (2.5, 2.5);
	
	\foreach \a in {4,...,0}
		\def \x { \a / 10}
		\filldraw[fill=white, draw=black] (0 + \x,0 + \x) rectangle (2 + \x,2 + \x);	
	
	%amount of modules		
	\draw[<->] (-0.1, 2.1) -- (0.9, 3.1) node[pos=0.5,sloped,above] {OFM Channels};
	\draw[<->] (-0.15, 0.0) -- (-0.15, 2) node[pos=0.5,sloped,above] {OFM Dim};
	\draw[<->] (0.0, -0.15) -- (2, -0.15) node[pos=0.5,sloped,below] {OFM Dim};

	%labels for modules
	\draw(1,1) node {OFM};

\end{tikzpicture}

\begin{tikzpicture}
	
	%modules
	\foreach \a in {10,...,6}
		\def \x { \a / 10}
		\filldraw[fill=white, draw=black] (0 + \x,0 + \x) rectangle (2 + \x,2 + \x);
		
	\foreach \a in {5,...,0}
		\def \x { \a / 10}
		\filldraw[fill=white, draw=black] (0 + \x,0 + \x) rectangle (2 + \x,2 + \x);	
	
	\foreach \a in {10,...,0}
		\def \x { \a / 10}
		\filldraw[fill=white, draw=red, line width = 0.4mm] (2 + \x,2 + \x) rectangle (1.8 + \x,1.8 + \x);
	
	%amount of modules		
	\draw[<->] (-0.1, 2.1) -- (0.9, 3.1) node[pos=0.5,sloped,above] {OFM Channels};
	\draw[<->] (-0.15, 0.0) -- (-0.15, 2) node[pos=0.5,sloped,above] {OFM Dim};
	\draw[<->] (0.0, -0.15) -- (2, -0.15) node[pos=0.5,sloped,below] {OFM Dim};

	%labels for modules
	\draw(1,1) node {OFM};

\end{tikzpicture}

}
\caption{Possible error models due to faulty PE, channel stuck@ (left) and same pixel stuck@ (right)} 
\label{fig:faulty PE}
\vspace{-0.5cm}
\end{figure}

In this work, we focus on \glspl{QNN} with different precisions for weights \textbf{W} and activations \textbf{A}, denoted as \textbf{W}\textit{w}\textbf{A}\textit{a}, where \textit{w} is the weights bit width, and \textit{a} is the activation bit widths. 
Different errors can appear in the output activation depending on the precision. For example in FINN for 1 bit activation, possible stuck@ values are \{0 (representing -1), 1\}. 
For 2 bits activation, since we are using symmetric quantization, the possible stuck@ values are \{-1, 0, 1\}.
% because merging the batch norm parameters and quantized activation function into the thresholding, requires flipping the weights whenever there is a negative value of gamma (in batch norm) and +2 cannot be represented in 2 bits. So,
\begin{comment}
\begin{figure}[htp]
\centering
\begin{subfigure}{.24\textwidth}
  \centering
  \includegraphics[width=3.3cm]{images/Picture1.png}
    \caption{Same pixel in all channels}
    \label{fig:FPE_pix}
\end{subfigure}%
\begin{subfigure}{.24\textwidth}
  \centering
  \includegraphics[width=3.5cm]{images/Picture2.png}
  \caption{Single \acrshort{ofm} channel}
  \label{fig: FPE_ch}
\end{subfigure}

\caption{Possible error models due to faulty \acrshort{pe}}
\label{fig:test}
\end{figure}
\end{comment}

\section{Related Work}
\label{sec:related work}

In the domain of self driving cars, recent works~\cite{jha2019kayotee, Understanding_error_propagation, 10.1145/3361566} have tried to assess the safety of \glspl{DNN}. 
If we look at the prior work done in making \glspl{DNN} error tolerant, most of the literature targets \glspl{MLP}, focusing mainly on stuck@0 faults~\cite{712162, 7862272, 714236, 5726611, 105414,duddu2019adversarial}. 
They propose different training methodologies to make \glspl{MLP} robust to errors.
%\citefull{712162} used min-max optimization with single node stuck@0 on \gls{MLP}.
\citefull{712162} formulated fault tolerance as a min-max optimization problem. They present two different methods to solve this min-max problem. The resulting \gls{NN} from both method though not always fault tolerant, exhibits acceptable degree of partial fault tolerance to the loss of single node (neuron) and its associated weights (representing stuck@0 error).
\citefull{7862272} includes the weighted sum of some special terms in the loss function (e.g. sum of squares of first/second derivatives of loss w.r.t weights and L2 norm of the weight vector) during training. They showed on 20 different datasets that resulting \gls{MLP} is robust to link-failures (stuck@0) and uncertainty in weights caused by additive or multiplicative noise.
\citefull{714236} train with stuck@0 faults injected during training and enhance the generalization ability of the \gls{MLP}. They study the relationship between fault tolerance and generalization and also show 6\% improvement in accuracy when training with error injection.
More recently, researchers have started looking into fault tolerance for \glspl{CNN}, such as the work by~\citefull{duddu2019adversarial} implements adversarial training for the feature extractor of \gls{CNN} and uses supervised training for the classifier to make it robust to stuck@0 errors only.
We are not aware of any works which attempt to train \glspl{QNN} to be robust to stuck@ errors, nor are we aware of any works which train \glspl{CNN} to be robust stuck@$y$ errors, where $y \neq 0$.

Similarly, there have only been few publications on the fault tolerance analysis of \glspl{QNN} with permanent faults exploiting the inherent features of the application. \citefull{FT_QNN} presented an error injection framework which provides a methodology to assess the error tolerance of \glspl{QNN} under a ingle channel stuck@ fault in hardware. 
They evaluated the error tolerance of \glspl{CNN} which were trained using \gls{SAT} and showed that convolutional layers are more sensitive to errors and causes a significant drop in accuracy as compared to \gls{FC} layers. 
As concluded from~\citefull{FT_QNN}, while giving comparable accuracy to floating point \glspl{CNN}, quantized \glspl{CNN} suffer from high error sensitivity.
As they are the preferred choice for deployment, their error tolerance is of particular importance. 

Before introducing the FAT methodology, we will provide a motivating example on why fault tolerance of \glspl{QNN} needs to be addressed for their deployment on embedded devices. The example is reproduced from~\citefull{FT_QNN} and shows the behaviour of \gls{SAT} under a particular error model and highlights the deficiency of \gls{SAT} which we will address using \gls{FAT}.

\subsection{Motivation}
\label{subsec:SATEval}
%\citefull{FT_QNN} evaluated the error tolerance of \glspl{CNN} trained with \gls{SAT} using Theano \cite{2016arXiv160502688short}. 
We trained the same \glspl{QNN} on CIFAR10 i.e. CNVW1A1, CNVW1A2 and CNVW2A2 in \gls{SAT} using the Brevitas library~\cite{brevitas} in PyTorch \cite{NEURIPS2019_9015}.
%Unlike~\citefull{FT_QNN} we did not use validation set, only training and test sets were used.
The \glspl{CNN} were evaluated using the error injection framework, using the channel stuck@ error model in which every channel in every layer is stuck at a time. The qualitative metric is the accuracy on the test set, which is computed in presence of the injected error.
The pseudo code for the channel stuck@ experiment is given in \cref{pseudocode}, for more details please refer to the work by \citefull{FT_QNN}.

\begin{algorithm}
\SetAlgoLined
 \For{every error value:}{
    \For{every layer:}{
        \For{every channel:}{
             stuck the channel @ error value and calculate accuracy on test set\;
        }
        list min and max accuracy observed for the layer\;
    }
 }
\caption{Error injection evaluation (channel stuck@)}
\label{pseudocode}
\end{algorithm}

\cref{table:SAT_cnvW1A1} shows the results for the error injection on CNVW1A1.
The network achieved higher accuracy than~\citefull{FT_QNN} but a  very high accuracy drop (down to 50.83\% for the worst case in layer 1) can be observed for \gls{SAT} under single channel stuck@. A drop of $>$30\% in accuracy might not acceptable for mission critical applications where lives or reputation of business/organization can be at stake.
%Similarly the results for CNVW1A2 and CNVW2A2 are listed in \cref{table:exp_sum}, which behaves similarly to the networks presented by~\citefull{FT_QNN} which were trained in Theano~\cite{2016arXiv160502688short}. 
This significant drop in accuracy under channel stuck@ faults indicates a possible deficiency of \gls{SAT} to incorporate fault tolerance in the \glspl{QNN}.

\begin{table}
\caption{SAT on CNVW1A1 with CIFAR10}\label{table:SAT_cnvW1A1}
\centering
	\begin{tabular}{|c|c|c|c|c|c|}
		\hline
		\multicolumn{2}{|c|}{\textbf{Error-free:} 84.57\%} & \multicolumn{2}{|c|}{\textbf{s@0 accuracy} [\%]} & \multicolumn{2}{|c|}{\textbf{s@1 accuracy} [\%]} \\ \hline
\textbf{Layer}  & \textbf{Channels}   & \textbf{min}     & \textbf{max}    & \textbf{min}    & \textbf{max}   \\ \hline
		0       & 64                  & 79.69            & 84.38           & 80.94           & 84.66          \\ \hline
		1       & 64                  & 79.81            & 84.78           &\textbf{50.83}   & 84.45          \\ \hline
		2       & 128                 & 76.33            & 84.67           & 78.46           & 84.66          \\ \hline
		3       & 128                 & 83.21            & 84.54           & 78.17           & 84.56          \\ \hline
		4       & 256                 & 82.63            & 84.62           & 82.76           & 84.71          \\ \hline
		5       & 256                 & 84.24            & 84.75           & 84.25           & 84.67          \\ \hline
		6       & 512                 & 84.40            & 84.66           & 84.38           & 84.69          \\ \hline
		7       & 512                 & 84.50            & 84.64           & 84.51           & 84.63          \\ \hline
	\end{tabular}
	
\end{table}

%The data listed in \cref{table:SAT_cnvW1A1} can also be visualized using a scatter plot, with minimum and maximum accuracy observed per each layer in the \gls{CNN}, on x and y axis respectively.
%In the scatter plot shown in \cref{fig:SAT_cnvW1A1}, the error-free accuracy is 84.57\% and the points from the \cref{table:SAT_cnvW1A1} are plotted.
For a more visual representation, in \cref{fig:SAT_cnvW1A1} we plot the data from \cref{table:SAT_cnvW1A1} as a scatter plot.
In \cref{fig:SAT_cnvW1A1} the point representing the minimum accuracy is the leftmost point.
The closer the minimum accuracy point is to the error-free point, the higher error tolerance the network exhibits.

In order to guarantee a high minimum accuracy under stuck@ errors, when implementing \glspl{QNN} trained with \gls{SAT}, one could apply generic techniques to make \gls{CNN} inference more reliable.
For example, hardware could be deployed with built-in redundancy, such as \gls{TMR} \cite{5392355}.
However, as the name suggests, this requires triplicating the \gls{DNN} hardware and further introduces an overhead  in the voting system which is used make the decision.
As such, standard \gls{TMR} introduces over a 200\% increase in compute and memory resources.

In this work we attempt to achieve reliable implementation of \gls{QNN} hardware accelerators, by exploiting inherent \gls{QNN} properties to decrease the overall hardware cost.
We utilise the inherent redundancy of \gls{QNN} themselves to make them more tolerant to errors.
Specifically, we incorporate stuck@ errors during \gls{QNN} training which produces highly error tolerant networks, without significantly affecting accuracy or introducing an increase in hardware cost.
This work relies on the error injection framework presented by~\citefull{FT_QNN} which enables fast characterization of the different training methodologies used in this work by means of \gls{FPGA} acceleration of \glspl{QNN}.
%It presents an error injection framework to evaluate the \acrshort{qnns} under a particular error model. We will be using the same error injection framework to evaluate our \acrshort{fat} trained \acrshort{cnns} and to compare them with the ones trained with SAT.

%The methodology proposed here is widely applicable w.r.t the works mentioned above and targets Quantized \acrshort{cnns} by enabling the user to introduce any error model and inject any possible errors probabilistically. 
%The resulting \acrshort{qnn} will exhibit higher error tolerance for the error model.

\pgfplotsset{
   every axis/.append style = {
                    label style={font=\small},
                    tick label style={font=\footnotesize} 
                }
}
\pgfplotstableread[col sep = comma]{images/data/SAT_CNVW1A1.csv}\saa

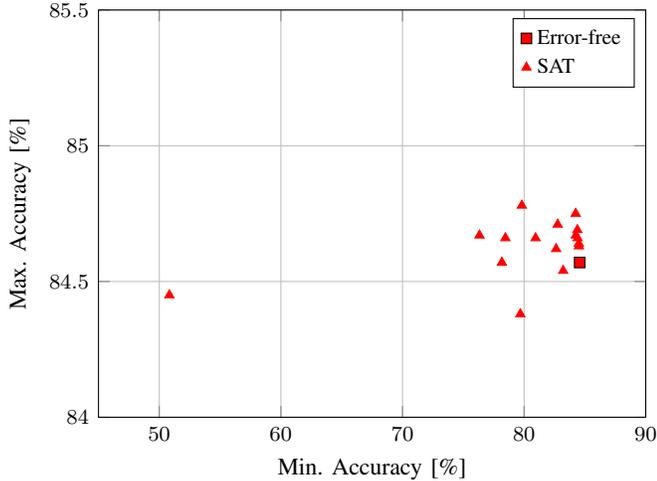
\begin{figure}[t]
\begin{tikzpicture}[scale=1]
\begin{axis}[
  grid =major,
  ylabel={Max. Accuracy [\%]},
  xlabel={Min. Accuracy [\%]},
  xmin=45, xmax=90,
  ymin=84, ymax=85.5,
  %legend style={draw=none,
  %at={(0.5,1.03)},
  %legend columns=5,
  %anchor=south},
  width=\columnwidth,
  height=7 cm,
  legend style={font=\footnotesize},
  legend style={at={(0.98,0.98)},anchor=north east},
  legend cell align={left}
]

\addlegendimage{only marks, mark=square*,black,style={solid, fill=red}}
\addlegendentry{Error-free};

\addlegendimage{only marks, mark=triangle*,red}
\addlegendentry{SAT};

\addplot[red, only marks, mark size=2pt, mark=triangle*] table[y = max_acc, x = min_acc] from \saa;

\addplot[black, only marks, mark size=2pt, mark=square*,style={solid, fill=red}] table[y = base, x = min_acc] from \saa;

\end{axis}
\end{tikzpicture}
\caption{Error-free and min./max. accuracy under single channel stuck@ error for CNVW1A1 trained with \gls{SAT} on CIFAR10}\label{fig:SAT_cnvW1A1}
\vspace{-0.5cm}
\end{figure}

\section{Fault Aware Training}
\label{sec:FAT}
\cref{fig:SAT_cnvW1A1} shows high variation among accuracies under single channel stuck@ error model.
The goal of \gls{FAT} is to train \glspl{QNN} such that they exhibit higher accuracy in presence of errors, thus decreasing the effects of faults within layer and among layers. 
%Note that, variance is not something which we try to minimize by adding it to the loss function. Rather, it is just a number which represents how well our trained \acrshort{cnn} is performing under a specific error model.
%Recall that the array of \acrshort{pes} in FINN, calculates entire \acrshort{ofm}, a faulty \acrshort{pe} in the dataflow affects the \acrshort{ofm} channels. 
The main idea is to inject errors (with a certain probability), which matches the error model of our inference hardware during training time. 
In order to implement the injection at training time, we developed an \textit{error injection layer} which is able to inject stuck@ errors in: channel(s), different pixels in different channels, and same pixel of all channels, depending on the chosen error model.
%The details of injection layer are described in \cref{subsec:EJ}.  

\subsection{Error Injection Layer}\label{subsec:EJ}
The error injection layer was developed in PyTorch~\cite{NEURIPS2019_9015} and it allows injecting errors with a particular error model during training.
The error injection layer receives a tensor as input in forward pass and injects the errors with a probability $p$, which is a hyper-parameter. 
As we chose to inject errors in the activation tensor to match our error model, $p$ represents the global probability of all possible errors to be injected (which depends on the activation precision). Thus each error value has a probability to be injected which is the division of $p$ over the number of possible errors.
%With the error injection layer, a particular error model can be included during training.
In the simplest case, the injection layer works as follows:
Given an $N$-dimensional input tensor, $\boldsymbol{\alpha} \in \mathbb{R}^{n_0 \times \cdots \times n_{N-1}}$, a random tensor, $\mathbf{r} \in \mathbb{R}^{n_0 \times \cdots \times n_{N-1}}$, the output, $\hat{\boldsymbol{\alpha}} \in \mathbb{R}^{n_0 \times \cdots \times n_{N-1}}$, of the injection layer is defined as follows:
%\cref{forward_pass,backward_pass}.
\begin{equation}\label{forward_pass}
    \hat{\alpha}_\mathbf{i} =%     
    \begin{cases}
        \epsilon_{0}\,\         & \mathrm{for}\ r_\mathbf{i} < p_{0} \\
        \epsilon_{1}\,\         & \mathrm{for}\ p_{0} \leq r_\mathbf{i} < p_{1} \\
        & \vdots \\
        \epsilon_{M-1}\,\       & \mathrm{for}\ p_{M-2} \leq r_\mathbf{i} < p_{M-1} \\
        \alpha_\mathbf{i}\,,\   & \mathrm{for}\ r_\mathbf{i} \geq p_{M-1}
    \end{cases}\ \ \mathrm{,}
\end{equation}
%where $\boldsymbol{\hat{\alpha}}$ is the output of the injection layer, $\boldsymbol{\alpha}$ represents the input tensor, $\boldsymbol{r_{d}} \sim \mathcal{U}[0,\,1)$ is a random tensor sampled from uniform distribution in every forward pass. 
%$d$ is the shape of the random tensor $\boldsymbol{r_{d}}$ which depends on the error model. For channel stuck@ model, it is $(batch size, channels)$, broadcasted over height and width of the incoming tensor. For same pixel in all channel stuck@ model it is $(batch size, height, width)$, broadcasted over all channels of the incoming tensor. 
%, and 
where $\mathbf{i} = (i_0, \cdots, i_{N-1})$ is a multi-index extracting a single element of an $N$-dimensional tensor, with $i_n \in \mathbb{Z}^+\ \forall\ n$,
$\alpha_\mathbf{i}$ is the $\mathbf{i}^{\mathrm{th}}$ element of $\boldsymbol{\alpha}$,
$\hat{\alpha}_{\mathbf{i}}$ is the $\mathbf{i}^{\mathrm{th}}$ element of $\hat{\boldsymbol{\alpha}}$,
$r_{\mathbf{i}} \sim U(0,1)$ is the $\mathbf{i}^{\mathrm{th}}$ element of $\mathbf{r}$,
$\epsilon_{m}$ is the $m^\mathrm{th}$ possible error value out of total $M$ activation values, where $M=2^b$, and $b$ is the bitwidth of the activation quantization scheme and
finally, $p_{j}$ is given by:
\begin{equation}\label{p_calc}
    p_{j} = \frac{p}{100}\frac{j}{M-1}\ \ \mathrm{,}
\end{equation}
where $j \in \mathbb{N}_0 \leq M-1$.

Similarly, given the loss gradients, $\frac{\partial L}{\partial \tilde{\boldsymbol{\alpha}}}  \in \mathbb{R}^{n_0 \times \cdots \times n_{N-1}}$,
the input gradients, $\frac{\partial L}{\partial \boldsymbol{\alpha}}  \in \mathbb{R}^{n_0 \times \cdots \times n_{N-1}}$, the backward pass of the injection layer is defined as follows:
\begin{equation}\label{backward_pass}
    \frac{\partial L}{\partial \alpha_\mathbf{i}} =%     
    \begin{cases}
        0\,,\  & \mathrm{for}\ r_{\mathbf{i}} < p \\
        \frac{\partial L}{\partial \hat{\alpha}_\mathbf{i}}\,,\  & \mathrm{for}\ r_{\mathbf{i}} \geq p
    \end{cases}\ \ \mathrm{,}
\end{equation}
where $\frac{\partial L}{\partial \hat{\alpha}_\mathbf{i}}$ is the $\mathbf{i}^{\mathrm{th}}$ element of $\frac{\partial L}{\partial \hat{\boldsymbol{\alpha}}}$ and
$\frac{\partial L}{\partial \alpha_\mathbf{i}}$ is the $\mathbf{i}^{\mathrm{th}}$ element of $\frac{\partial L}{\partial \boldsymbol{\alpha}}$.
%The injection layer supports injecting multiple types of stuck@ errors, which define the shape of $r_d$. 
Multiple fault models are accommodated within a single injection layer through the definition of $\mathbf{r}$ and a \textit{broadcasting} function\footnote{https://numpy.org/doc/stable/user/theory.broadcasting.html}, $\mathcal{B}_{\boldsymbol{\alpha}}$.
For example, the broadcasting function $\mathcal{B}_{\boldsymbol{\alpha}}(\cdot)$ expands an input tensor by replicating values across every singleton dimension of the input tensor so that it has the same shape as $\boldsymbol{\alpha}$.
For image processing, where the input is a 4D tensor, $\boldsymbol{\alpha} \in \mathbb{R}^{b \times c \times h \times w}$, where $b$\,/\,$c$\,/\,$h$\,/\,$w$ refers to batch size / channels / height / width respectively.
In order to inject channel stuck@ errors, $\mathbf{r}$ is defined as:
\begin{equation}
\label{eq:random_broadcastable_channel}
\mathbf{r} = \mathcal{B}_{\boldsymbol{\alpha}} (\mathbf{r}_c)\ \ \mathrm{,}
\end{equation}
where $\mathbf{r}_c \in \mathbb{R}^{b \times c \times 1 \times 1}$, with each element of $\mathbf{r}_c$ is drawn from $U(0,1)$.
The broadcasting function replicates the values of $\mathbf{r}_c$ in the singleton dimensions (height and width, in this example), resulting in identical values within each channel within each batch.
When this value of $\mathbf{r}$ is used in \cref{forward_pass}, the result that each channel is either stuck at a particular value or not.
Similarly, for the pixel stuck@ model we have:
\begin{equation}
\label{eq:random_broadcastable_pixel}
\mathbf{r} = \mathcal{B}_{\boldsymbol{\alpha}} (\mathbf{r}_p)\ \ \mathrm{,}
\end{equation}
where $\mathbf{r}_p \in \mathbb{R}^{b \times 1 \times h \times w}$, with each element of $\mathbf{r}_p$ is drawn from $U(0,1)$.

\begin{figure}
\centering
\resizebox{\columnwidth}{!}{
\begin{tikzpicture}
    
    \foreach \a in {1,...,6}
    {
    	\def \x { \a*.5}
    	% upper box
    	\filldraw[fill=white, draw=black] (0.3 , 1+\x) rectangle (1, 1.5+\x);
    	\draw(0.65, 1.25+\x) node {\dataA[\a,1]};
        % lower box
        \filldraw[fill=white, draw=black] (0.3 , -2.5+\x) rectangle (1, -2+\x);
        
        \ifnum \a=1
            \def \col {red} 
        \else 
            \ifnum \a=2
                \def \col {red}
            \else
                \ifnum \a=5
                    \def \col {red}
                \else
                    \def \col {black}
                \fi
            \fi
        \fi
        \draw[color=\col](0.65, -2.25+\x) node {\dataB[\a,1]};
	}
	
	% lines for a and der
	\draw[->, color=black, line width=0.4mm] (1.3, 3) -- (4.8, 1.7) node[pos=0.5,sloped,above] {$\boldsymbol{\alpha}$};
	\draw[<-, color=red, line width=0.4mm] (1.3, -0.5) -- (4.8, 0.8) node[pos=0.5,sloped,below] {$\dfrac{\partial L}{\partial \boldsymbol{\alpha}}$};
	
	% injection module
	\filldraw[fill=white, draw=red, line width = 0.4mm] (5,-0.5) rectangle (6, 3);	
    \draw[color=red](5.5,1.3) node {\begin{turn}{90}Injection Layer\end{turn}};
    \draw(5.5,-0.8) node {status = \textit{enable}};
    \draw(5.5,-1.3) node {\textit{p} = \textbf{50\%}};
    
    % lines for a and der
 	\draw[->, color=red, line width = 0.4mm] (6.2, 1.7) -- (9.7, 3) node[pos=0.5,sloped,above] {$\hat{\boldsymbol{\alpha}}$};	
	\draw[<-, color=black, line width = 0.4mm] (6.2, 0.8) -- (9.7, -0.5) node[pos=0.5,sloped,below] {$\dfrac{\partial L}{\partial \hat{\boldsymbol{\alpha}}}$};

    \foreach \a in {1,...,6}
    {
    	\def \x { \a*.5}
    	% upper box
    	\filldraw[fill=white, draw=black] (10 , 1+\x) rectangle (10.7, 1.5+\x);
        
        \ifnum \a=1
            \def \col {red} 
        \else 
            \ifnum \a=2
                \def \col {red}
            \else
                \ifnum \a=5
                    \def \col {red}
                \else
                    \def \col {black}
                \fi
            \fi
        \fi
        
    	\draw[color=\col](10.35, 1.25+\x) node {\dataC[\a,1]};
        % lower box
        \filldraw[fill=white, draw=black] (10 , -2.5+\x) rectangle (10.7, -2+\x);
        \draw(10.35, -2.25+\x) node {\dataD[\a,1]};
	}

\end{tikzpicture}
}
\caption{Forward and backward pass of injection layer} \label{fig:backprop}
\vspace{-0.5cm}
\end{figure}
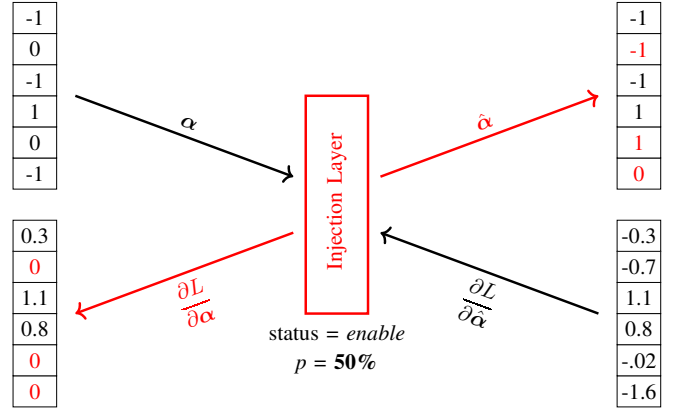

The probability $p_{i}$ for each of the error value is calculated according to the \cref{p_calc}. For example, given a 5\% probability $p$ of injection in CNVW1A1, each of the possible error values \{-1, 1\} has a probability of 2.5\%. 
%During backward pass, the injection layer zero outs the gradients at the indices where the error was injected during forward pass. 
\cref{fig:backprop} shows an example of forward and backward pass for an \gls{FC} layer with \textbf{A}=2 and possible error values are \{-1, 0, 1\}. 
Whenever the status variable is set to \textit{enable}, the layer is actively injecting errors. On the other hand, setting the status to \textit{disable} will make the injection layer transparent (passing the tensor without injecting errors).
The error injection probability in \cref{fig:backprop} is set at $p$=50\%, hence every activation value in the tensor (abbreviated as $\boldsymbol{\alpha}$ and $\boldsymbol{\hat{\alpha}}$) has 50\% probability of being replaced by error.
During backward pass, gradients of the loss with respect to the activations (abbreviated as $\frac{\partial L}{\partial \boldsymbol{\alpha}}$) at the indices at which error was injected, are replaced with zero. 
This is performed because a small change in the corresponding input activations (which were erroneous after the injection) did not effect the loss value.
 %In the remaining of the work, we will be considering the channel stuck@ error model. 

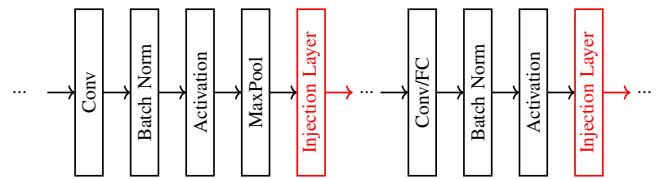
\begin{figure}[!b]
\centering
\resizebox{\columnwidth}{!}{
\begin{tikzpicture}
	
	% lines in
	\draw[line width = 0.3mm] (-1, 1.5) node{...};
	\draw[->, color=black, line width=0.3mm] (-0.5, 1.5) -- (0, 1.5);
	\filldraw[fill=white, draw=black, line width = 0.3mm] (0,0) rectangle (0.5, 3);	
    \draw[color=black](0.25,1.5) node {\begin{turn}{90}Conv\end{turn}};
    \draw[->, color=black, line width=0.3mm] (0.5, 1.5) -- (1, 1.5);
    
	\filldraw[fill=white, draw=black, line width = 0.3mm] (1,0) rectangle (1.5, 3);	
    \draw[color=black](1.25,1.5) node {\begin{turn}{90}Batch Norm\end{turn}};
    \draw[->, color=black, line width=0.3mm] (1.5, 1.5) -- (2, 1.5);

	\filldraw[fill=white, draw=black, line width = 0.3mm] (2,0) rectangle (2.5, 3);	
    \draw[color=black](2.25,1.5) node {\begin{turn}{90}Activation\end{turn}};
    \draw[->, color=black, line width=0.3mm] (2.5, 1.5) -- (3, 1.5);
    
	\filldraw[fill=white, draw=black, line width = 0.3mm] (3,0) rectangle (3.5, 3);	
    \draw[color=black](3.25,1.5) node {\begin{turn}{90}MaxPool\end{turn}};
    \draw[->, color=black, line width=0.3mm] (3.5, 1.5) -- (4, 1.5);
    
	% injection module
	\filldraw[fill=white, draw=red, line width = 0.3mm] (4,0) rectangle (4.5, 3);	
    \draw[color=red](4.25,1.5) node {\begin{turn}{90}Injection Layer\end{turn}};
 	\draw[->, color=red, line width = 0.3mm] (4.5, 1.5) -- (5, 1.5);
 	
 	\draw[line width = 0.3mm] (5.25, 1.5) node{...};
 
	\draw[->, color=black, line width=0.3mm] (5.5, 1.5) -- (6, 1.5);
	\filldraw[fill=white, draw=black, line width = 0.3mm] (6,0) rectangle (6.5, 3);	
    \draw[color=black](6.25,1.5) node {\begin{turn}{90}Conv/FC\end{turn}};
    \draw[->, color=black, line width=0.3mm] (6.5, 1.5) -- (7, 1.5);
	
	\filldraw[fill=white, draw=black, line width = 0.3mm] (7,0) rectangle (7.5, 3);	
    \draw[color=black](7.25,1.5) node {\begin{turn}{90}Batch Norm\end{turn}};
    \draw[->, color=black, line width=0.3mm] (7.5, 1.5) -- (8, 1.5);

	\filldraw[fill=white, draw=black, line width = 0.3mm] (8,0) rectangle (8.5, 3);	
    \draw[color=black](8.25,1.5) node {\begin{turn}{90}Activation\end{turn}};
    \draw[->, color=black, line width=0.3mm] (8.5, 1.5) -- (9, 1.5);

	% injection module
	\filldraw[fill=white, draw=red, line width = 0.3mm] (9,0) rectangle (9.5, 3);	
    \draw[color=red](9.25,1.5) node {\begin{turn}{90}Injection Layer\end{turn}};
 	\draw[->, color=red, line width = 0.3mm] (9.5, 1.5) -- (10, 1.5);
 	
 	\draw[line width = 0.3mm] (10.25, 1.5) node{...};
 	
\end{tikzpicture}
}
\caption{Injection layer in CNV example} \label{fig:ins}
\vspace{-0.5cm}
\end{figure}

\cref{fig:ins} shows the insertion of the injection layer in the \gls{CNN} which is usually composed of repeating blocks i.e. convolution/\gls{FC}, batch-norm, activation, and sometimes includes pooling. As the injection layer is designed to inject errors in activation tensor, it should be included after activation layer or a pooling layer.

%Every injection layer has a \textit{status} and probability of injection \textbf{p}. The status can be set to 1) \textit{Enable} - to inject errors during training according to the target error model, 2) \textit{Disable} - to pass the tensor as it is (for error-free evaluation) and 3) \textit{Eval} - to perform GPU evaluation of the error injection (used for bigger models like ImageNet). 
%In our experiments we did not include the injection layer in the \gls{FC} layers as it is shown to be error tolerant~\cite{8875314}. We will also verify our assumption by evaluating the \gls{FC} layer under error injection that not injecting in the \gls{FC} layers does not cause the minimum accuracy

%\tikzsetnextfilename{folding}

\def\data{-1 0 1 -1 0 -1}
\readarray\data\dataA[6,1]

\def\data{0 0 0.8 1.1 0 0.3}
\readarray\data\dataB[6,1]

\def\data{0 1 1 -1 -1 -1}
\readarray\data\dataC[6,1]

\def\data{-1.6 -.02 0.8 1.1 -0.7 -0.3}
\readarray\data\dataD[6,1]

\subsection{Error Injection vs Dropout}\label{subsec:EJVD}

The behaviour of the injection layer is conceptually similar to Dropout~\cite{hinton2012improving} and Dropout2D~\cite{tompson2015efficient}.
In this subsection, we describe the similarities and differences between our proposed error injection layer and dropout variants.
Specifically in this work, we refer to the dropout implementations in PyTorch~\cite{NEURIPS2019_9015}.

Dropout was first introduced by \citefull{hinton2012improving} as a method of reducing overfitting (i.e., improving model generalisation) by preventing the co-adaption of activations.
The Dropout layer takes an $N$-dimensional tensor as an input, $\boldsymbol{\alpha} \in \mathbb{R}^{n_0 \times \cdots \times n_{N-1}}$,
and produces a output tensor, $\tilde{\boldsymbol{\alpha}} \in \mathbb{R}^{n_0 \times \cdots \times n_{N-1}}$, with the same shape to the input, but with some values replaced with 0.
I.e., some values of $\boldsymbol{\alpha}$ are ``dropped out''.
To normalise the vector, the remaining values are scaled by $1/(1-p)$,
where $p$ is the probability of the neuron being set to zero.
Explicitly, the forward path result, $\tilde{\boldsymbol{\alpha}}$, of Dropout is calculated from $\boldsymbol{\alpha}$ and a random tensor $\mathbf{r} \in \mathbb{R}^{n_0 \times \cdots \times n_{N-1}}$ as follows:
\begin{equation}
    \label{eq:dropout_forward_function}
    \tilde{\alpha}_{\mathbf{i}} =%     
    \begin{cases}
        0\,,\  & \mathrm{for}\ r_{\mathbf{i}} < p \\
        \frac{1}{1-p}\alpha_\mathbf{i}\,,\  & \mathrm{for}\ r_{\mathbf{i}} \geq p
    \end{cases}\ \ \mathrm{,}
\end{equation}
where $\mathbf{i} = (i_0, \cdots, i_{N-1})$ is a multi-index, with $i_n \in \mathbb{Z}^+\ \forall\ n$,
$\alpha_\mathbf{i}$ is the $\mathbf{i}^{\mathrm{th}}$ element of $\boldsymbol{\alpha}$,
$\tilde{\alpha}_{\mathbf{i}}$ is the $\mathbf{i}^{\mathrm{th}}$ element of $\tilde{\boldsymbol{\alpha}}$ and
$r_{\mathbf{i}} \sim U(0,1)$ is the $\mathbf{i}^{\mathrm{th}}$ element of $\mathbf{r}$.
In a standard case, $N=2$ where $n_0$ is the batch size and $n_1$ is the number of neurons in the layer preceding the Dropout layer.
Similarly in the backward path, given loss gradients, $\frac{\partial L}{\partial \tilde{\boldsymbol{\alpha}}}  \in \mathbb{R}^{n_0 \times \cdots \times n_{N-1}}$,
the input gradients, $\frac{\partial L}{\partial \boldsymbol{\alpha}}  \in \mathbb{R}^{n_0 \times \cdots \times n_{N-1}}$, are calculated as follows:
\begin{equation}
    \label{eq:dropout_backward_function}
    \frac{\partial L}{\partial \alpha_\mathbf{i}} =%     
    \begin{cases}
        0\,,\  & \mathrm{for}\ r_{\mathbf{i}} < p \\
        \frac{1}{1-p}\frac{\partial L}{\partial \tilde{\alpha}_\mathbf{i}}\,,\  & \mathrm{for}\ r_{\mathbf{i}} \geq p
    \end{cases}\ \ \mathrm{,}
\end{equation}
where $\frac{\partial L}{\partial \tilde{\alpha}_\mathbf{i}}$ is the $\mathbf{i}^{\mathrm{th}}$ element of $\frac{\partial L}{\partial \tilde{\boldsymbol{\alpha}}}$ and
$\frac{\partial L}{\partial \alpha_\mathbf{i}}$ is the $\mathbf{i}^{\mathrm{th}}$ element of $\frac{\partial L}{\partial \boldsymbol{\alpha}}$.

An extension to Dropout, Dropout2D, was introduced by \citefull{tompson2015efficient} to extend Dropout to improve its effectiveness in 2D convolutional layers of neural networks.
Dropout2d operates on 4-dimensional tensors, with the dimensions, $[n_0, n_1, n_2, n_3]$, usually corresponding to batch size, channels, height and width respectively.
Instead of dropping out individual values in the input tensor, Dropout2D drops out and entire channel to zero if the value of $r_\mathbf{i} < p$.

The injection layer varies from dropout in the following ways:
\begin{itemize}
    \item it supports injecting all possible fault values into the activations, while dropout only inserts zeros;
    \item dropout applies scaling to both the outputs and the loss gradients on uncorrupted values, while the error injection layer does not; and
    \item the error injection layer supports better slicing of the incoming tensor, including stuck@ pixel modes while the dropout variants do not.
\end{itemize}

\subsection{FAT methodology}
\label{subsec:fat}
After the injection layer is ready to be included in the training of the \glspl{QNN}, we performed experiments with different configuration of injection layer, error models, and precisions.  
The probability $p$ of injecting errors is an additional training hyper-parameter which can be set to a fixed value before starting the training process or it can be modified during the training.
The value of $p$ needs to be tuned and depends according to the data set, \gls{CNN} topology and precision.
We applied the FAT methodology using the injection layer described in \cref{subsec:EJ} with multiple $p$ values and configurations of the injection layer. In order to characterize the methodology, the first set-up explored (method 1) was training \glspl{CNN} with:
\begin{itemize}
    \item Enabling all the injection layers;  
    \item Fixed probability $p$ of injection throughout training.
\end{itemize}
The second approach (method 2) is based on enabling only one of the injection layers in each training epoch (see \cref{pseudocode_XFAT}). 
\begin{algorithm}
\SetAlgoLined
 \For{every epoch:}{
    enable one random injection layer and disable others \newline
    \For{every batch:}{
         forward and backward pass to train model
    }
 }
\caption{FAT Method 2}
\label{pseudocode_XFAT}
\end{algorithm}

\section{Experiments and Results}
\label{sec:experimental}
We trained CNVW1A1, CNVW1A2 CNVW1A3, CNVW1A4, and CNVW2A2 with \gls{SAT} and \gls{FAT}, targeting channel stuck@ error model on CIFAR10, GTSRB, SVHN as well as ImageNet (using DoReFa-Net by~\citefull{zhou2016dorefanet}). 
For same pixel in all channel stuck@, we experimented with CNVW1A1, CNVW1A2 and CNVW2A2 on CIFAR10 only.
For \gls{SAT} and \gls{FAT} the training hyper-parameters were kept the same i.e., 1000 epochs, ADAM optimizer with initial learning rate of 0.02 which was halved every 40 epochs, weight decay set to 0, batch-size of 100 and squared hinge loss. 
Training was performed on NVIDIA P6000 GPU. The error injection evaluation for CNVW1A1, CNVW1A2 and CNVW2A2 was performed on a Xilinx Zynq UltraScale+ MPSOC ZCU104~\cite{zcu104}, Whereas for ImageNet it was performed on multiple GPUs.
%To start with, we trained 3 networks the probability to 5\%, 2.5\% and 1.5\%. The scatter plot for CNVW1A1 of \gls{SAT} vs 5\%, 2.5\% and 1.5\% \gls{FAT} are given in \cref{fig:FAT_CNVW1A1_comp}. 

\subsection{SAT vs FAT Method 1}
In method 1 we enabled all injection layers in our \gls{CNN} with a fixed probability $p$. \cref{fig:FAT_CNVW1A1_comp} shows a comparison between error injection results of SAT CNVW1A1 (same as in \cref{fig:SAT_cnvW1A1}) and 3 CNVW1A1 trained with FAT (method 1) on CIFAR10 with different probabilities of error injection. For \gls{FAT}, the injection probabilities are 5\%, 2.5\% and 1.5\% and the error model used is channel stuck@.
Note, that the \gls{SAT} error-free accuracy is occluded by the \gls{FAT}($p$=1\%) error-free accuracy in \cref{fig:FAT_CNVW1A1_comp}.
%Similarly, we trained from scratch CNVW1A2 and CNVW2A2 () for different probability values of error injection. 
It can be noticed from the scatter plots that training with \gls{FAT} makes the \gls{CNN} more resilient to channel stuck@ errors, given that for all the 3 networks trained with \gls{FAT} the minimum accuracy in presence of a single channels stuck@ ($x$-axis in \cref{fig:FAT_CNVW1A1_comp}) is much closer to the error-free accuracy and it shows a huge improvement. 
Nonetheless, it can also be noticed how the error-free accuracy in some cases is lower than the \gls{SAT} one. 
As expected, by increasing the error injection probability, the \gls{CNN} increases the robustness against channel stuck@ errors, while decreasing the error-free accuracy. 
Similarly, by decreasing the value $p$ the error-free accuracy gets closer to the original trained with \gls{SAT}. 
% Why are we referencing the table? are those results with the second methodology?
%This is most likely due to the regularization effect that the injection layer has on the trained network, decreasing the over-fitting. %Nick please have a look at this sentence! 
%Low variance with low error-free accuracy on high probability of error injection is because injecting after every layer, injects more error than need and it grows. 

\pgfplotsset{
   every axis/.append style = {
                    label style={font=\small},
                    tick label style={font=\footnotesize} 
                }
}
\pgfplotstableread[col sep = comma]{images/data/FAT_CNVW1A1_comp.csv}\saa

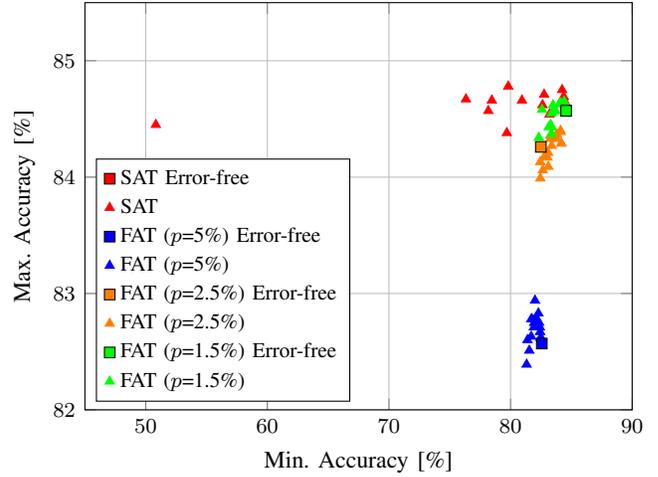
\begin{figure}
\begin{tikzpicture}[scale=1]
\begin{axis}[
  grid =major,
  ylabel={Max. Accuracy [\%]},
  xlabel={Min. Accuracy [\%]},
  xmin=45, xmax=90,
  ymin=82, ymax=85.5,
  %legend style={draw=none,
  %at={(0.5,1.03)},
  %legend columns=5,
  %anchor=south},
  width=\columnwidth,
  height=7 cm,
  legend style={font=\footnotesize},
  legend style={at={(0.02,0.02)},anchor=south west},
  legend cell align={left}
]

%\addlegendimage{only marks, mark=square,black}
%\addlegendentry{Error-free};

\addlegendimage{only marks, mark=square*,black,style={solid, fill=red}}
\addlegendentry{SAT Error-free};

\addlegendimage{only marks, mark=triangle*,red}
\addlegendentry{SAT};

\addlegendimage{only marks, mark=square*,black,style={solid, fill=blue}}
\addlegendentry{FAT ($p$=5\%) Error-free};

\addlegendimage{only marks, mark=triangle*,blue}
\addlegendentry{FAT ($p$=5\%)};

\addlegendimage{only marks, mark=square*,black,style={solid, fill=orange}}
\addlegendentry{FAT ($p$=2.5\%) Error-free};

\addlegendimage{only marks, mark=triangle*,orange}
\addlegendentry{FAT ($p$=2.5\%)};

\addlegendimage{only marks, mark=square*,black,style={solid, fill=green}}
\addlegendentry{FAT ($p$=1.5\%) Error-free};

\addlegendimage{only marks, mark=triangle*,green}
\addlegendentry{FAT ($p$=1.5\%)};

\addplot[red, only marks, mark size=2pt, mark=triangle*] table[y = SAT_max_acc, x = min_acc] from \saa;

\addplot[blue, only marks, mark size=2pt, mark=triangle*] table[y = FAT5_max_acc, x = min_acc] from \saa;

\addplot[orange, only marks, mark size=2pt, mark=triangle*] table[y = 
FAT25_max_acc, x = min_acc] from \saa;

\addplot[green, only marks, mark size=2pt, mark=triangle*] table[y = FAT15_max_acc, x = min_acc] from \saa;

\addplot[black, only marks, mark size=2pt, mark=square*, style={solid, fill=red}] table[y = SAT_base, x = min_acc] from \saa;

\addplot[black, only marks, mark size=2pt, mark=square*, style={solid, fill=blue}] table[y = FAT5_base, x = min_acc] from \saa;

\addplot[black, only marks, mark size=2pt, mark=square*, style={solid, fill=orange}] table[y = FAT25_base, x = min_acc] from \saa;

\addplot[black, only marks, mark size=2pt, mark=square*, style={solid, fill=green}] table[y = FAT15_base, x = min_acc] from \saa;

\end{axis}
\end{tikzpicture}
\caption{Error-free and min./max. accuracy under single channel stuck@ error for SAT vs FAT (method 1) on CNVW1A1 with CIFAR10}\label{fig:FAT_CNVW1A1_comp}
\vspace{-0.5cm}
\end{figure}

\subsection{SAT vs FAT Method 2}
In method 2, only one of the injection layers in the network is kept enabled at a time, with a fixed probability $p$, while all other injection layers are kept disabled.
For this configuration we also trained the networks with different probabilities of error injection i.e. from 1\% to 20\%. 
The results for FAT method 2 on CNVW1A1 with 5\%, 10\% and 15\% probability and channel stuck@ error model are given in \cref{fig:XFAT_CNVW1A1}. 
For method 2 (enabling one layer at a time) it can be observed that both 5\% and 10\% provides better results than the method 1 (with all the injection layers enabled concurrently).
First, in all cases with different $p$ values we always notice a similar improvement as method 1 in presence of single channel stuck@ error i.e. minimum accuracy always being higher than 80\%. 
Second, an improvement in error-free accuracy for low probabilities of error injection i.e. 5\% and 10\% as compared to SAT can be clearly observed from \cref{fig:XFAT_CNVW1A1}.
In \cref{table:exp_sum}, a similar trend was observed for higher precisions.
For all of the networks tested, we achieved higher minimum accuracy under single channel stuck@ as well as higher error-free accuracy. 
%\cref{table:exp_sum} shows the actual numbers and results for CNVW1A2 and CNVW2A2 results for single injection per epoch with different probabilities of error injection. 
Method 2 ($p$=5\% and 10\%) provides better error-free accuracies than method 1 ($p$=5\%) which is 85.07\% and 84.75\% respectively as compared to 82.57\% and the min accuracy is also better for method 2 ($p$=10\%) i.e. 82.43\% as compared to 81.32\%.
Interestingly for higher probabilities of injection e.g. 15\%, the error-free accuracy for \gls{FAT} method 2 was lower than \gls{SAT}. 
Finally, it should also be noted that training with \gls{FAT} comes at no additional convergence time or number of epochs.
Since FAT (method 2) outperformed method 1 in most of our experiments, we performed extended experiments on FAT method 2. Furthermore, unless specified otherwise any reference to FAT from this section onwards refers to FAT (method 2).
In \cref{table:exp_sum}, we also show results for experiments extended to new datasets, i.e. GTSRB and SVHN (on CNVW1A1), as well as changing the \gls{CNN} topology i.e. ImageNet (trained using DoReFa-Net by ~\citefull{zhou2016dorefanet} for which the possible error values in the activations are \{0, 0.333, 0.666, 1\}. On all datasets, minimum and error-free accuracies observed with FAT were better than SAT. \cref{fig:XFAT_CNVW1A2_Imagenet} shows the results on ImageNet with injection probability of 5\% and 2.5\%. Note, the ``Ref.'' column in \cref{table:exp_sum} refers to the error-free accuracy of equivalent \gls{SAT} models from the available literature.

\pgfplotsset{
   every axis/.append style = {
                    label style={font=\small},
                    tick label style={font=\footnotesize} 
                }
}
\pgfplotstableread[col sep = comma]{images/data/XFAT_CNVW1A1.csv}\saa

\begin{figure}
\begin{tikzpicture}[scale=1]
\begin{axis}[
  grid =major,
  ylabel={Max. Accuracy [\%]},
  xlabel={Min. Accuracy [\%]},
  xmin=45, xmax=90,
  ymin=82, ymax=85.5,
%  legend style={draw=none,
%  at={(0.5,1.03)},
%  legend columns=5,
%  anchor=south},
  width=\columnwidth,
  height=7 cm,
  legend style={font=\footnotesize},
  legend style={at={(0.02,0.02)},anchor=south west},
  legend cell align={left}
]

%\addlegendimage{only marks, mark=square,black}
%\addlegendentry{Error-free};

\addlegendimage{only marks, mark=square*,black,style={solid, fill=red}}
\addlegendentry{SAT Error-free};

\addlegendimage{only marks, mark=triangle*,red}
\addlegendentry{SAT};

\addlegendimage{only marks, mark=square*,black,style={solid, fill=green}}
\addlegendentry{FAT ($p$=5\%) Error-free};

\addlegendimage{only marks, mark=triangle*,green}
\addlegendentry{FAT ($p$=5\%)};

% \addlegendimage{only marks, mark=triangle*,blue}
% \addlegendentry{FAT (p=5\%)};

% \addlegendimage{only marks, mark=square*,blue}
% \addlegendentry{FAT (p=5\%) Error-free};

\addlegendimage{only marks, mark=square*,black,style={solid, fill=cyan}}
\addlegendentry{FAT ($p$=10\%) Error-free};

\addlegendimage{only marks, mark=triangle*,cyan}
\addlegendentry{FAT ($p$=10\%)};

\addlegendimage{only marks, mark=square*,black,style={solid, fill=brown}}
\addlegendentry{FAT ($p$=15\%) Error-free};

\addlegendimage{only marks, mark=triangle*,brown}
\addlegendentry{FAT ($p$=15\%)};

\addplot[red, only marks, mark size=2pt, mark=triangle*] table[y = SAT_max_acc, x = min_acc] from \saa;

\addplot[green, only marks, mark size=2pt, mark=triangle*] table[y = XFAT5_max_acc, x = min_acc] from \saa;

% \addplot[blue, only marks, mark size=2pt, mark=triangle*] table[y = FAT5_max_acc, x = min_acc] from \saa;

\addplot[cyan, only marks, mark size=2pt, mark=triangle*] table[y = 
XFAT10_max_acc, x = min_acc] from \saa;

\addplot[brown, only marks, mark size=2pt, mark=triangle*] table[y = XFAT15_max_acc, x = min_acc] from \saa;

\addplot[black, only marks, mark size=2pt, mark=square*,style={solid, fill=red}] table[y = base, x = min_acc] from \saa;

\addplot[black, only marks, mark size=2pt, mark=square*,style={solid, fill=green}] table[y = XFAT5_base, x = min_acc] from \saa;

\addplot[black, only marks, mark size=2pt, mark=square*,style={solid, fill=cyan}] table[y = XFAT10_base, x = min_acc] from \saa;

\addplot[black, only marks, mark size=2pt, mark=square*,style={solid, fill=brown}] table[y = XFAT15_base, x = min_acc] from \saa;

\end{axis}
\end{tikzpicture}
\caption{Error-free and min./max. accuracy under single channel stuck@ error for SAT vs FAT (method 2) on CNVW1A1 with CIFAR10}\label{fig:XFAT_CNVW1A1}
\vspace{-0.5cm}
\end{figure}
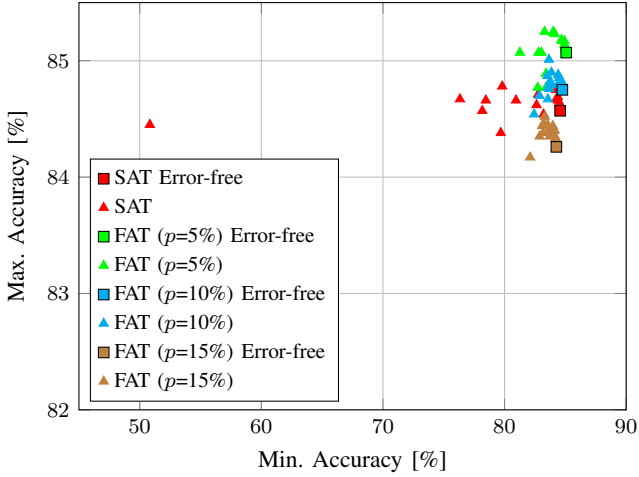
\pgfplotsset{
   every axis/.append style = {
                    label style={font=\small},
                    tick label style={font=\footnotesize} 
                }
}
\pgfplotstableread[col sep = comma]{images/data/XFAT_CNVW1A2_Imagenet.csv}\saa

%\vspace{-0.5cm}
\begin{figure}
\begin{tikzpicture}[scale=1]
\begin{axis}[
  grid =major,
  ylabel={Max. Accuracy [\%]},
  xlabel={Min. Accuracy [\%]},
  xmin=20, xmax=60,
  ymin=52, ymax=55.5,
%  legend style={draw=none,
%  at={(0.5,1.03)},
%  legend columns=5,
%  anchor=south},
  width=\columnwidth,
  height=7 cm,
  legend style={font=\footnotesize},
  legend style={at={(0.02,0.02)},anchor=south west},
  legend cell align={left}
]

%\addlegendimage{only marks, mark=square,black}
%\addlegendentry{Error-free};

\addlegendimage{only marks, mark=square*,black,style={solid, fill=red}}
\addlegendentry{SAT Error-free};

\addlegendimage{only marks, mark=triangle*,red}
\addlegendentry{SAT};

%\addlegendimage{only marks, mark=triangle*,orange}
%\addlegendentry{XFAT (p=10 per batch\%)};

\addlegendimage{only marks, mark=square*,black,style={solid, fill=blue}}
\addlegendentry{FAT ($p$=5\%) Error-free};

\addlegendimage{only marks, mark=triangle*,blue}
\addlegendentry{FAT ($p$=5\%)};

\addlegendimage{only marks, mark=square*,black,style={solid, fill=green}}
\addlegendentry{FAT ($p$=2.5\%) Error-free};

\addlegendimage{only marks, mark=triangle*,green}
\addlegendentry{FAT ($p$=2.5\%)};

\addplot[red, only marks, mark size=2pt, mark=triangle*] table[y = SAT_max_acc, x = min_acc] from \saa;

%\addplot[orange, only marks, mark size=2pt, mark=triangle*] table[y = %XFAT10_max_acc, x = min_acc] from \saa;

\addplot[blue, only marks, mark size=2pt, mark=triangle*] table[y = XFAT5_max_acc, x = min_acc] from \saa;

\addplot[green, only marks, mark size=2pt, mark=triangle*] table[y = XFAT25_max_acc, x = min_acc] from \saa;

\addplot[black, only marks, mark size=2pt, mark=square*,style={solid, fill=red}] table[y = base, x = min_acc] from \saa;

\addplot[black, only marks, mark size=2pt, mark=square*,style={solid, fill=blue}] table[y = XFAT5_base, x = min_acc] from \saa;

\addplot[black, only marks, mark size=2pt, mark=square*,style={solid, fill=green}] table[y = XFAT25_base, x = min_acc] from \saa;

\end{axis}
\end{tikzpicture}
\caption{Error-free and min./max. accuracy under single channel stuck@ error for SAT vs FAT (method 2) on DoReFa-Net with Imagenet}\label{fig:XFAT_CNVW1A2_Imagenet}
\vspace{-0.5cm}
\end{figure}
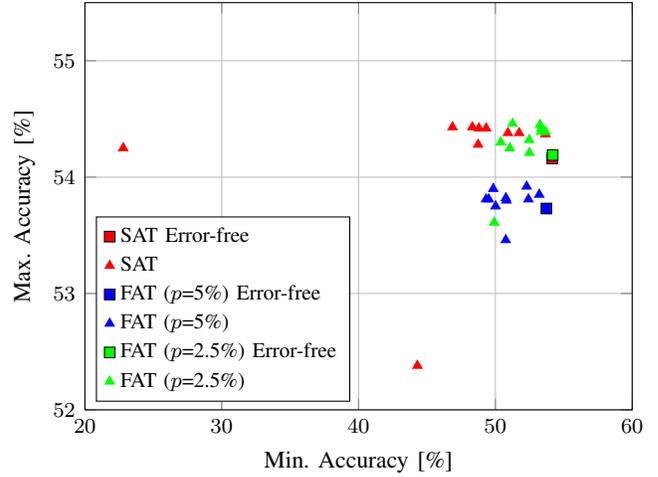

\begin{table*}
\caption{Experiments Summary: SAT vs FAT (channel s@)}\label{table:exp_sum}
% \centering
	\begin{tabular}{|c|c|c|c|c|c|c|c|c|c|c|c|c|c|}
	\hline
\multirow{2}{*}{Network} & \multirow{2}{*}{Dataset}   & \multicolumn{3}{|c|}{SAT} &\multicolumn{4}{|c|}{FAT}  & \multirow{2}{*}{Ref. [\%]} \\ \cline{3-5} \cline{6-9}

& & Error-free [\%]  & Min Acc. [\%] & Variance & Error-free [\%]   & Min Acc. [\%] & Variance    & Prob. ($p$\%) & \\ \hline

CNVW1A1 &CIFAR10 &84.57 &50.83 &1.211  &\textbf{85.07} &81.26 &0.150 &5.0 &79.22~\cite{finn-r} \\ \hline
CNVW1A1 &CIFAR10 &- &- &-  &84.75 &\textbf{82.43} &0.085 &10.0 &- \\ \hline
CNVW1A1 &CIFAR10 &- &- &- &84.25 &82.09 &0.052 &15.0 &- \\ \hline
CNVW1A2 &CIFAR10 &87.46 &64.41 &2.612 &\textbf{88.31} &\textbf{84.78} &0.153 &5.0 &82.66~\cite{finn-r} \\ \hline
CNVW1A2 &CIFAR10 &- &- &- &87.72 &84.77 &0.092 &10.0 &- \\ \hline
CNVW1A3 &CIFAR10 &89.21  &80.06 &2.331 &\textbf{89.73} &\textbf{87.90} &0.089 &5.0 &- \\ \hline
CNVW1A4 &CIFAR10 &89.13  &80.69 &1.960 &\textbf{89.74} &\textbf{88.45} &0.074 &5.0 &- \\ \hline

CNVW2A2 &CIFAR10 &88.25 &67.98 &2.508 &\textbf{88.96} &86.07 &0.057 &5.0 &84.29~\cite{finn-r} \\ \hline
CNVW2A2 &CIFAR10 &- &- &- &88.60 &\textbf{86.66} &0.043 &10.0 &- \\ \hline
CNVW1A1 &SVHN    &95.73 &94.15 &0.04 &\textbf{95.95} &\textbf{95.14} &0.01 &5.0 &94.90~\cite{finn-r}\\ \hline	
CNVW1A1 &GTSRB   &98.11 &95.89 &0.06 &\textbf{98.70} &\textbf{98.00} &0.01 &5.0 &98.08~\cite{finn-r}\\ \hline
DoReFa-Net &ImageNet&54.16 &22.80 &2.93 &\textbf{54.19} &\textbf{49.92} &0.78 &2.5 &49.80~\cite{zhou2016dorefanet}\\ \hline	
	\end{tabular}
	
\end{table*}

\subsection{Selective Channel Replication with FAT}

One of the hardware solutions presented by~\citefull{FT_QNN} is selective channel replication, in which rather than triplicating a whole \gls{CNN} (\gls{TMR}), only selective portion of the network is triplicated. 
The methodology consist of characterizing all channels of each layer based on their impact on accuracy under channel stuck@ error model, sorting them from most to least critical, and triplicating the most critical ones.
The amount of channels to triplicate depends on the worst case accuracy in presence of single channel stuck@ error that is required to be guaranteed in a system.
This provides bigger design space in the hardware cost vs. worst case accuracy trade-off. 
For example in \cref{fig:pareto} $x$-axis is the normalized hardware cost of a fully unrolled model where each \gls{MAC} operation in the neural network has its own \gls{PE}, and the $y$-axis is the worst case error in the presence of single ch stuck@ (in this case, error means the percentage of images that are misclassified, i.e. 100 - minimum accuracy observed).
On the plot, the right most point of any of the curve represents triplicating all of the channels, hence it costs the maximum hardware but guarantees the same worst case error (under single channel stuck@) as the error free.
Whereas the left most point of the curve represents no triplication at all and hence low hardware cost but higher worst case error. Similarly middle points represents triplicating a few channels and the corresponding worst case error.
From \cref{fig:pareto} we can clearly see that \gls{FAT} offers better trade-offs providing all the pareto dominating solutions in the design space. 
For example, in order to guarantee 13.84\% worst case error for CNVW2A2, \gls{SAT} requires triplicating 278 channels whereas \gls{FAT} requires triplicating 2 channels only.

\pgfplotsset{
   every axis/.append style = {
                    label style={font=\small},
                    tick label style={font=\footnotesize} 
                }
}
\pgfplotstableread[col sep = comma]{images/data/XTMR.csv}\saa
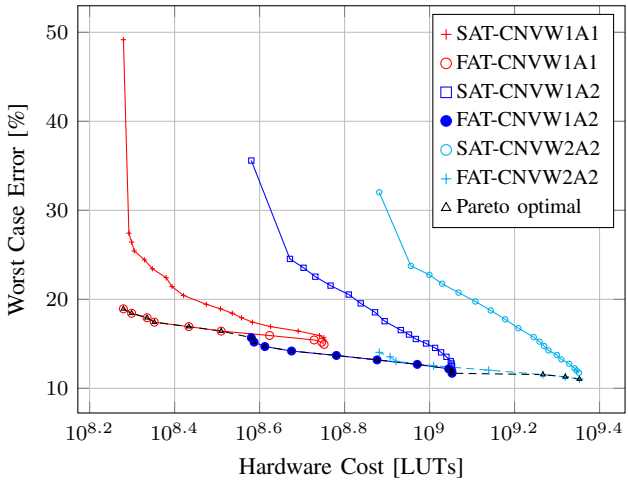
\begin{figure}[t]
\begin{tikzpicture}[scale=1]
\begin{semilogxaxis}[
  grid =major,
  ylabel={Worst Case Error [\%]},
  xlabel={Hardware Cost [LUTs]},
%  legend style={draw=none,
%  at={(0.5,1.03)},
%  legend columns=4,
%  anchor=south},
  width=\columnwidth,
  height=7 cm,
  legend style={font=\footnotesize},
  legend style={at={(0.98,0.98)},anchor=north east},
  legend cell align={left}
]
\addlegendimage{only marks, mark=+,red}
\addlegendentry{SAT-CNVW1A1};

\addlegendimage{only marks, mark=o,red}
\addlegendentry{FAT-CNVW1A1};

\addlegendimage{only marks, mark=square,blue}
\addlegendentry{SAT-CNVW1A2};

\addlegendimage{only marks, mark=*,blue}
\addlegendentry{FAT-CNVW1A2};

\addlegendimage{only marks, mark=o,cyan}
\addlegendentry{SAT-CNVW2A2};

\addlegendimage{only marks, mark=+,cyan}
\addlegendentry{FAT-CNVW2A2};

\addlegendimage{only marks, mark=triangle,black}
\addlegendentry{Pareto optimal};

\addplot +[red, sharp plot, mark size=1pt, mark=+ ] 
table [y = SAT-cnvW1A1, x = Hardware Cost] {\saa};

\addplot +[red, sharp plot, mark size=1.5pt, mark=o] 
table [y = XFAT-cnvW1A1, x = Hardware Cost] {\saa};

\addplot +[blue, sharp plot, mark size=1pt, mark=square] 
table [y = SAT-cnvW1A2, x = Hardware Cost] {\saa};

\addplot +[blue, sharp plot, mark size=1.5pt, mark=*] 
table [y = XFAT-cnvW1A2, x = Hardware Cost] {\saa};

\addplot +[cyan, sharp plot, mark size=1pt, mark=o] 
table [y = SAT-cnvW2A2, x = Hardware Cost] {\saa};

\addplot +[cyan, sharp plot, mark size=1.5pt, mark=+] 
table [y = XFAT-cnvW2A2, x = Hardware Cost] {\saa};

\addplot +[ black, sharp plot, mark size=1pt, mark=triangle ] 
table [y = Pareto, x = Hardware Cost] {\saa};

\end{semilogxaxis}
\end{tikzpicture}
\caption{Pareto frontier of worst-case test error vs. hardware cost with selective channel replication}\label{fig:pareto}
\vspace{-0.5cm}
\end{figure}

\subsection{FAT vs Dropout2D} %Please Nick have a look at this section

In order to compare with FAT, we trained the same networks with Dropout2D.
For the Dropout2D experiments, we replicated FAT (method 1), except each error injection layer is replaced with a Dropout2D layer.

\cref{fig:DROPOUT2D_CNVW1A1} shows the results for CNVW1A1 in which an improved minimum accuracy can be observed with both 5\% and 2.5\% probability of Dropout2D. For example, the minimum accuracy observed under single channel stuck@ is 80.14\% as compared to 81.26\% of FAT. \cref{table:exp_sum_FAT_D2D} shows a comparison between Dropout2D and FAT in terms of error-free, minimum accuracy and overall variance in the accuracies observed.
As shown in the table, Dropout2D also performs very well in terms of error-free accuracy, outperforming \gls{FAT} in 2 of the 5 networks tested.
In terms of minimum accuracy, surprisingly, Dropout2D outperforms \gls{FAT} in 4 of the 5 networks tested.
Although in the average case, there is little difference between the accuracy of \gls{FAT} and Dropout2D for both error-free and minimum accuracy.
There are a few exceptions, notably networks CNVW1A1, CNVW1A3 and CNVW1A4 have approximately a 1\% accuracy difference in the minimum accuracy.
Both techniques can be considered very effective in improving the minimum accuracy under a single channel stuck@ error model.
A surprising observation, is that even though Dropout2D only injects 0-values into tensors it appears robust to stuck@ errors associated with larger integer values, such as 7 and 15, as shown by the results of the cnvW1A3 and cnvW1A4 networks.
This will be investigated in further future works.

As Dropout2D achieves similar results to FAT for the channel stuck@ error model in the experiments performed on \glspl{QNN}, we performed additional experiments on different error model. 
\cref{table:exp_sum_SAT_FAT_D2D_pixel} shows a detailed comparison among SAT, FAT and Dropout2D on the same pixel stuck@ error model (shown in the right side of \cref{fig:faulty PE}).
The networks were evaluated with same pixel stuck@ injection, in which we iterate over all pixels in the $(height, width)$ dimension of the tensor and stuck that pixel for all channels to a particular error value.
%This type of error model could be appropriate for an accelerator with different scheduling to FINN, for example, one in which each PE processes a single pixel in the output across multiple channels.
For FAT and Dropout2D, we experimented with different probabilities and the best results achieved in term of minimum accuracies are listed in \cref{table:exp_sum_SAT_FAT_D2D_pixel}.
Similar to the single channel stuck@ model, the \glspl{QNN} trained with SAT exhibit a high drop in accuracy in presence of single pixel stuck@ error. For example, CNVW1A2 suffers from a drop of $>$20\% accuracy, from 87.46\% in the error-free case down to 64.93\%. 
The networks trained with FAT method 2 while injecting errors according to the pixel stuck@ error model shows higher error-free accuracy, as well as very high tolerance to single pixel stuck@ errors as compared to SAT.
CNVW1A1, CNVW1A2, and CNVW2A2 trained with FAT shows a drop of 1.43\%, 1.51\%, and 1.13\% respectively, showing a similar trend to the experiments for channel stuck@ listed in \cref{table:exp_sum_FAT_D2D}. 
As compared to SAT, Dropout2D shows increase in accuracy as well, in the presence of single error, while in all tested cases FAT outperformed Dropout2D.
We believe that the regularization effect provided by Dropout2D is very promising in terms of tolerance for only some error models. 
Nonetheless, the results in \cref{table:exp_sum_SAT_FAT_D2D_pixel} show that during training, it is important to adopt the error model which better represents possible hardware faults in the target accelerator. FAT provides this additional level of flexibility.
It is possible, that with more flexible control over the dropout pattern in the output tensor, Dropout will also perform well for this error model.

%Note that, in case of \textbf{A}=2, the error-free accuracy is better for Dropout2D as compared to FAT.
%\cref{fig:DROPOUT2D_CNVW1A1} and \cref{table:exp_sum_FAT_D2D} shows the results for dropout vs FAT. 
%also acts as a regularizer and algorithmically has very similar properties to \gls{FAT}. 
%In our experiments dropout2d turns out also to be helpful in making a CNN fault tolerant and the results are given in \cref{table:exp_sum_FAT_D2D}.
%In our experiments, \gls{FAT} outperforms dropout2d, offering better min accuracy and better error-free i.e. making a \gls{QNN} resilient to hardware stuck@ faults without any trade-offs. 

\pgfplotsset{
   every axis/.append style = {
                    label style={font=\small},
                    tick label style={font=\footnotesize} 
                }
}
\pgfplotstableread[col sep = comma]{images/data/DROPOUT2D_CNVW1A1.csv}\saa

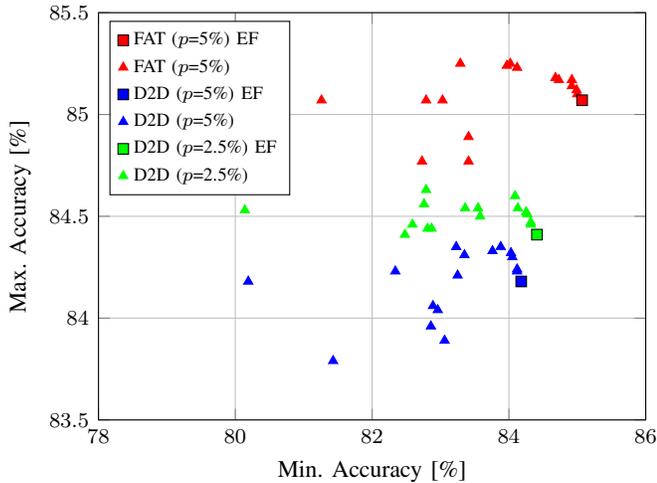
\begin{figure}[t]
\begin{tikzpicture}[scale=1]
\begin{axis}[
  grid =major,
  ylabel={Max. Accuracy [\%]},
  xlabel={Min. Accuracy [\%]},
  xmin=78, xmax=86,
  ymin=83.5, ymax=85.5,
%  legend style={draw=none,
%  at={(0.5,1.03)},
%  legend columns=5,
%  anchor=south},
  width=\columnwidth,
  height=7 cm,
  legend style={font=\scriptsize},
  legend style={at={(0.02,0.98)},anchor=north west},
  legend cell align={left}
]

%\addlegendimage{only marks, mark=square,black}
%\addlegendentry{Error-free};

\addlegendimage{only marks, mark=square*,black,style={solid, fill=red}}
\addlegendentry{FAT ($p$=5\%) EF};

\addlegendimage{only marks, mark=triangle*,red}
\addlegendentry{FAT ($p$=5\%)};

\addlegendimage{only marks, mark=square*,black,style={solid, fill=blue}}
\addlegendentry{D2D ($p$=5\%) EF};

\addlegendimage{only marks, mark=triangle*, blue}
\addlegendentry{D2D ($p$=5\%)};

\addlegendimage{only marks, mark=square*,black,style={solid, fill=green}}
\addlegendentry{D2D ($p$=2.5\%) EF};

\addlegendimage{only marks, mark=triangle*, green}
\addlegendentry{D2D ($p$=2.5\%)};

\addplot[red, only marks, mark size=2pt, mark=triangle*] table[y = SAT_max_acc, x = min_acc] from \saa;

\addplot[blue, only marks, mark size=2pt, mark=triangle*] table[y = 
FAT5_max_acc, x = min_acc] from \saa;

\addplot[green, only marks, mark size=2pt, mark=triangle*] table[y = FAT25_max_acc, x = min_acc] from \saa;

\addplot[black, only marks, mark size=2pt, mark=square*,style={solid, fill=red}] table[y = base, x = min_acc] from \saa;

\addplot[black, only marks, mark size=2pt, mark=square*,style={solid, fill=blue}] table[y = FAT5_base, x = min_acc] from \saa;

\addplot[black, only marks, mark size=2pt, mark=square*,style={solid, fill=green}] table[y = FAT25_base, x = min_acc] from \saa;

\end{axis}
\end{tikzpicture}
\caption{FAT vs Dropout2D (D2D) on CNVW1A1 with CIFAR10 of both error-free (EF) and with a single channel stuck@ error}\label{fig:DROPOUT2D_CNVW1A1}
\vspace{-0.5cm}
\end{figure}

\begin{table*}[t]
\caption{Experiments Summary: FAT vs Dropout2D (channel s@)}\label{table:exp_sum_FAT_D2D}
% \centering
	\begin{tabular}{|c|c|c|c|c|c|c|c|c|c|c|c|c|c|}
	\hline
\multirow{2}{*}{Network} & \multirow{2}{*}{Dataset}   & \multicolumn{4}{|c|}{FAT} &\multicolumn{4}{|c|}{Dropout2D}  \\ \cline{3-10}

& & Error-free [\%] & Min Acc. [\%] & Variance & Prob. ($p$\%) & Error-free [\%] & Min Acc. [\%] & Variance & Prob. ($p$\%)  \\ \hline

CNVW1A1 &CIFAR10 &\textbf{85.07} &\textbf{81.26} &0.150 &5.0 &84.41 &80.14 &0.132 &2.5  \\ \hline
CNVW1A2 &CIFAR10 &88.31 &84.78 &0.153 &5.0 &\textbf{88.36} &\textbf{85.07} &0.113 &2.5  \\ \hline
CNVW1A3 &CIFAR10 &\textbf{89.56} &87.27 &0.089     &2.5 &89.50 &\textbf{88.38} &0.073     &2.5  \\ \hline
CNVW1A4 &CIFAR10 &\textbf{89.41} &87.86 &0.074     &2.5 &89.36 &\textbf{88.60} &0.079     &2.5  \\ \hline
CNVW2A2 &CIFAR10 &88.96 &86.07 &0.057 &5.0 &\textbf{89.47} &\textbf{86.33} &0.089 &2.5 \\ \hline
	\end{tabular}
	
\end{table*}

\begin{table*}[t]
\caption{Experiments Summary: SAT vs FAT vs Dropout2D (same pixel s@)}\label{table:exp_sum_SAT_FAT_D2D_pixel}
% \centering
\resizebox{\columnwidth*2}{!}{
	\begin{tabular}{|c|c|c|c|c|c|c|c|c|c|}
	\hline
\multirow{2}{*}{Network} & \multirow{2}{*}{Dataset} & \multicolumn{2}{|c|}{SAT} & \multicolumn{3}{|c|}{FAT} &\multicolumn{3}{|c|}{Dropout2d} \\ \cline{3-10}

& & Error-free [\%] & Min Acc. [\%] & Prob. ($p$\%) & Error-free [\%] & Min Acc. [\%] & Prob. ($p$\%) & Error-free [\%] & Min Acc. [\%] \\ \hline

CNVW1A1 &CIFAR10 & 84.57 & 79.83 & 2.5 & \textbf{85.03} & \textbf{83.60} & 2.5  & 84.41 & 81.14 \\ \hline
CNVW1A2 &CIFAR10 & 87.46 & 64.93 & 2.5 & 88.06 & \textbf{86.55} & 10.0 & \textbf{88.50} & 82.48 \\ \hline
% CNVW1A3 &CIFAR10 & 89.20 & 77.06 & 10.0 & 89.23 & 80.21 & 2.5  & 89.02 & 85.97  \\ \hline
% CNVW1A4 &CIFAR10 & 89.13 & 77.06 & 15.0 & 89.32 & 81.04 & 2.5  & 89.45 & 86.08\\ \hline
CNVW2A2 &CIFAR10 & 88.25 & 71.34 & 2.5 & \textbf{89.13} & \textbf{88.00} & 10.0 & 87.70 & 83.71 \\ \hline
	\end{tabular}
	}
\end{table*}

\section{Observations, Discussion and Suggestions}
\label{sec:observations}
It can be seen from the results in \cref{sec:experimental} that the \gls{FAT} methodology helps in training error tolerant \glspl{QNN}.
The error injection during training also acts as a regularizer and prevents overfitting, thus increasing the accuracy of the model. 
%In most of the previous work which targeted error injection, to make \glspl{MLP} fault tolerant to stuck@0 faults, a trade off in accuracy to some extent can be observed, as reflected by method 1.
When using method 1 with the higher probability of error injection i.e. $>$5\% very low variance in accuracies can be achieved but at the expense of lower error-free accuracy. 
Depending upon the application in which \glspl{CNN} is going to be deployed, if low variance among accuracies in presence of faults is the primary concern then method 1 is more suitable. 
Whereas if the high error-free accuracy is of higher priority then method 2 is recommended.
In our experiments, we found that values of $p$ in the range of 1-10\% resulting in the best error-free and worst-case accuracy.
Enabling one injection layer per epoch (method 2) and training from scratch is recommended to get a better minimum accuracy and lower overall variance in worst-case accuracy.

\section{Conclusion}
\label{sec:conclusion}

In this work we presented a new method for training an efficient error tolerant \glspl{QNN} called \gls{FAT}.
We showed that training with \gls{FAT} yields highly accurate and error tolerant \glspl{QNN} at no additional accuracy and convergence cost.
To the best of our knowledge \gls{FAT} outperforms all existing work targeting stuck@0 faults. In addition, we showed that training \glspl{QNN} with \gls{FAT} can also improve the error-free accuracy, suggesting that \gls{FAT} may behave like a regularizer, to improve generalization of \glspl{QNN} to the test set.
We also showed that a variant of Dropout, Dropout2D, can be used to significantly improve the tolerance of \glspl{QNN} to stuck@ faults, which provides similar performance to \gls{FAT} to for single channel stuck@ errors.
However, we also showed the importance of injecting errors during training which matches the associated fault model, in particular, showing that \gls{FAT} outperforms Dropout2D in the same pixel stuck@ case.
We also showed that \gls{FAT} offers an improved hardware cost vs. worst case error trade off when compensating for errors in hardware.
We are planning to extend our investigation and explore \gls{FAT} for different error models, higher precision, alternative layer types and alternative architectures.
Furthermore, in this work we applied \gls{FAT} to popular benchmarks, in future works we plan to apply these techniques to applications with strict accuracy requirements under hardware faults and to more complex training regimes, such as reinforcement learning and generative adversarial networks.
Lastly, we also plan to investigate ways to further scale the error injection evaluation campaign, in order to allow more complex network topologies to be explored.
%The future work includes studying multiple channel stuck@ and also targeting the loopback architecture, as well as extending the experiments to higher precisions. 
%Additionally, we would like to extend this idea not only to \glspl{CNN}, but also to other \glspl{DNN}.

%\bibliographystyle{ieee/IEEEtran}
\bibliography{ieee/IEEEexample}

% Generated by IEEEtranN.bst, version: 1.14 (2015/08/26)
\begin{thebibliography}{39}
\providecommand{\natexlab}[1]{#1}
\providecommand{\url}[1]{#1}
\csname url@samestyle\endcsname
\providecommand{\newblock}{\relax}
\providecommand{\bibinfo}[2]{#2}
\providecommand{\BIBentrySTDinterwordspacing}{\spaceskip=0pt\relax}
\providecommand{\BIBentryALTinterwordstretchfactor}{4}
\providecommand{\BIBentryALTinterwordspacing}{\spaceskip=\fontdimen2\font plus
\BIBentryALTinterwordstretchfactor\fontdimen3\font minus
  \fontdimen4\font\relax}
\providecommand{\BIBforeignlanguage}[2]{{%
\expandafter\ifx\csname l@#1\endcsname\relax
\typeout{** WARNING: IEEEtranN.bst: No hyphenation pattern has been}%
\typeout{** loaded for the language `#1'. Using the pattern for}%
\typeout{** the default language instead.}%
\else
\language=\csname l@#1\endcsname
\fi
#2}}
\providecommand{\BIBdecl}{\relax}
\BIBdecl

\bibitem[Krizhevsky et~al.(2012)Krizhevsky, Sutskever, and
  Hinton]{Krizhevsky:2012:ICD:2999134.2999257}
A.~Krizhevsky, I.~Sutskever, and G.~E. Hinton, ``Imagenet classification with
  deep convolutional neural networks,'' ser. NIPS'12, 2012, pp. 1097--1105.

\bibitem[{Redmon} et~al.(2016){Redmon}, {Divvala}, {Girshick}, and
  {Farhadi}]{redmon2015look}
J.~{Redmon}, S.~{Divvala}, R.~{Girshick}, and A.~{Farhadi}, ``You only look
  once: Unified, real-time object detection,'' in \emph{2016 IEEE CVPR}, 2016,
  pp. 779--788.

\bibitem[{Badrinarayanan} et~al.(2017){Badrinarayanan}, {Kendall}, and
  {Cipolla}]{badrinarayanan2015segnet}
V.~{Badrinarayanan}, A.~{Kendall}, and R.~{Cipolla}, ``{SegNet}: A deep
  convolutional encoder-decoder architecture for image segmentation,''
  \emph{IEEE TPAMI}, vol.~39, no.~12, pp. 2481--2495, 2017.

\bibitem[Ronneberger et~al.()Ronneberger, Fischer, and
  Brox]{ronneberger2015unet}
O.~Ronneberger, P.~Fischer, and T.~Brox, ``{U-Net}: Convolutional networks for
  biomedical image segmentation,'' in \emph{MICCAI 2015}, pp. 234--241.

\bibitem[He et~al.(2015)He, Zhang, Ren, and Sun]{he2015delving}
K.~He, X.~Zhang, S.~Ren, and J.~Sun, ``Delving deep into rectifiers: Surpassing
  human-level performance on imagenet classification,'' in \emph{ICCV}, 2015,
  pp. 1026--1034.

\bibitem[{Gysel} et~al.(2018){Gysel}, {Pimentel}, {Motamedi}, and
  {Ghiasi}]{gysel2016ristretto}
P.~{Gysel}, J.~{Pimentel}, M.~{Motamedi}, and S.~{Ghiasi}, ``Ristretto: A
  framework for empirical study of resource-efficient inference in
  convolutional neural networks,'' \emph{IEEE TNNLS}, 2018.

\bibitem[Courbariaux et~al.(2016)Courbariaux, Hubara, Soudry, El-Yaniv, and
  Bengio]{binary_net}
M.~Courbariaux, I.~Hubara, D.~Soudry, R.~El-Yaniv, and Y.~Bengio, ``Binarized
  neural networks: Training deep neural networks with weights and activations
  constrained to+ 1 or-1,'' \emph{arXiv preprint arXiv:1602.02830}, 2016.

\bibitem[Rastegari et~al.()Rastegari, Ordonez, Redmon, and
  Farhadi]{rastegari2016xnornet}
M.~Rastegari, V.~Ordonez, J.~Redmon, and A.~Farhadi, ``{XNOR-Net}: Imagenet
  classification using binary convolutional neural networks,'' in \emph{ECCV
  2016}.

\bibitem[Zhou et~al.(2016)Zhou, Wu, Ni, Zhou, Wen, and Zou]{zhou2016dorefanet}
S.~Zhou, Y.~Wu, Z.~Ni, X.~Zhou, H.~Wen, and Y.~Zou, ``{DoReFa-Net}: Training
  low bitwidth convolutional neural networks with low bitwidth gradients,''
  \emph{arXiv preprint arXiv:1606.06160}, 2016.

\bibitem[Cai et~al.(2017)Cai, He, Sun, and Vasconcelos]{cai2017deep}
Z.~Cai, X.~He, J.~Sun, and N.~Vasconcelos, ``Deep learning with low precision
  by half-wave gaussian quantization,'' pp. 5918--5926, 2017.

\bibitem[Han et~al.(2015)Han, Mao, and Dally]{han2015deep}
S.~Han, H.~Mao, and W.~J. Dally, ``Deep compression: Compressing deep neural
  networks with pruning, trained quantization and huffman coding,'' \emph{arXiv
  preprint arXiv:1510.00149}, 2015.

\bibitem[Howard et~al.(2017)Howard, Zhu, Chen, Kalenichenko, Wang, Weyand,
  Andreetto, and Adam]{MobileNet}
A.~G. Howard, M.~Zhu, B.~Chen, D.~Kalenichenko, W.~Wang, T.~Weyand,
  M.~Andreetto, and H.~Adam, ``Mobilenets: Efficient convolutional neural
  networks for mobile vision applications,'' vol. abs/1704.04861, 2017.

\bibitem[Wu et~al.(2017)Wu, Wan, Yue, Jin, Zhao, Golmant, Gholaminejad,
  Gonzalez, and Keutzer]{Shiftnet}
B.~Wu, A.~Wan, X.~Yue, P.~H. Jin, S.~Zhao, N.~Golmant, A.~Gholaminejad,
  J.~Gonzalez, and K.~Keutzer, ``Shift: {A} zero flop, zero parameter
  alternative to spatial convolutions,'' \emph{CoRR}, vol. abs/1711.08141,
  2017.

\bibitem[Fraser et~al.(2017)Fraser, Umuroglu, Gambardella, Blott, Leong, Jahre,
  and Vissers]{scalingBNN}
N.~J. Fraser, Y.~Umuroglu, G.~Gambardella, M.~Blott, P.~Leong, M.~Jahre, and
  K.~Vissers, ``Scaling binarized neural networks on reconfigurable logic,''
  ser. PARMA-DITAM ’17.\hskip 1em plus 0.5em minus 0.4em\relax New York, NY,
  USA: Association for Computing Machinery, 2017, p. 25–30.

\bibitem[Tripathi et~al.(2017)Tripathi, Dane, Kang, Bhaskaran, and
  Nguyen]{DBLP:journals/corr/TripathiDKBN17}
S.~Tripathi, G.~Dane, B.~Kang, V.~Bhaskaran, and T.~Q. Nguyen, ``Lcdet:
  Low-complexity fully-convolutional neural networks for object detection in
  embedded systems,'' \emph{CoRR}, vol. abs/1705.05922, 2017.

\bibitem[Umuroglu et~al.()Umuroglu, Fraser, Gambardella, Blott, Leong, Jahre,
  and Vissers]{finn}
Y.~Umuroglu, N.~J. Fraser, G.~Gambardella, M.~Blott, P.~Leong, M.~Jahre, and
  K.~Vissers, ``Finn: A framework for fast, scalable binarized neural network
  inference,'' in \emph{FPGA-17}.\hskip 1em plus 0.5em minus 0.4em\relax ACM,
  pp. 65--74.

\bibitem[Carlier et~al.(1996)Carlier, Coindoz, Deneuville, Garbellini, and
  Altavilla]{RAMS}
S.~Carlier, M.~Coindoz, L.~Deneuville, L.~Garbellini, and A.~Altavilla,
  ``Evaluation of reliability, availability, maintainability and safety
  requirements for manned space vehicles with extended on-orbit stay time,''
  \emph{Acta Astronautica}, vol.~38, no.~2, pp. 115 -- 123, 1996.

\bibitem[Goble and Brombacher(1999)]{FMEDA}
W.~Goble and A.~Brombacher, ``Using a failure modes, effects and diagnostic
  analysis (fmeda) to measure diagnostic coverage in programmable electronic
  systems,'' \emph{Reliability Engineering \& System Safety}, pp. 145 -- 148,
  1999.

\bibitem[{Gambardella} et~al.(2019){Gambardella}, {Kappauf}, {Blott},
  {Doehring}, {Kumm}, {Zipf}, and {Vissers}]{FT_QNN}
G.~{Gambardella}, J.~{Kappauf}, M.~{Blott}, C.~{Doehring}, M.~{Kumm},
  P.~{Zipf}, and K.~{Vissers}, ``Efficient error-tolerant quantized neural
  network accelerators,'' in \emph{DFT 19}, Oct. 2019.

\bibitem[Khoshavi et~al.(2020)Khoshavi, Broyles, and Bi]{SEU_BNN}
N.~Khoshavi, C.~Broyles, and Y.~Bi, ``A survey on impact of transient faults on
  bnn inference accelerators,'' \emph{arXiv preprint arXiv:2004.05915}, 2020.

\bibitem[Tompson et~al.(2015)Tompson, Goroshin, Jain, LeCun, and
  Bregler]{tompson2015efficient}
J.~Tompson, R.~Goroshin, A.~Jain, Y.~LeCun, and C.~Bregler, ``Efficient object
  localization using convolutional networks,'' in \emph{CVPR}, 2015.

\bibitem[Mittal(2018)]{Mittal2018}
S.~Mittal, ``A survey of fpga-based accelerators for convolutional neural
  networks,'' \emph{Neural Computing and Applications}, Oct 2018.

\bibitem[Venieris et~al.(2018)Venieris, Kouris, and Bouganis]{vkb18}
S.~I. Venieris, A.~Kouris, and C.-S. Bouganis, ``{Toolflows for Mapping
  Convolutional Neural Networks on FPGAs: A Survey and Future Directions},''
  \emph{ACM Computing Surveys (CSUR)}, vol.~51, no.~3, pp. 56--39, Jul. 2018.

\bibitem[Blott et~al.(2018)Blott, Preusser, Fraser, Gambardella, O'Brien, and
  Umuroglu]{finn-r}
M.~Blott, T.~Preusser, N.~Fraser, G.~Gambardella, K.~O'Brien, and Y.~Umuroglu,
  ``{FINN-R}: An end-to-end deep-learning framework for fast exploration of
  quantized neural networks,'' \emph{ACM TRETS}, 2018.

\bibitem[BNN()]{BNN-PYNQ-REPO}
\BIBentryALTinterwordspacing
 [Online]. Available: \url{https://github.com/Xilinx/BNN-PYNQ/}
\BIBentrySTDinterwordspacing

\bibitem[Jha et~al.(2019)Jha, Tsai, Hari, Sullivan, Kalbarczyk, Keckler, and
  Iyer]{jha2019kayotee}
S.~Jha, T.~Tsai, S.~Hari, M.~Sullivan, Z.~Kalbarczyk, S.~W. Keckler, and R.~K.
  Iyer, ``Kayotee: A fault injection-based system to assess the safety and
  reliability of autonomous vehicles to faults and errors,'' \emph{arXiv
  preprint arXiv:1907.01024}, 2019.

\bibitem[Li et~al.(2017)Li, Hari, Sullivan, Tsai, Pattabiraman, Emer, and
  Keckler]{Understanding_error_propagation}
G.~Li, S.~K.~S. Hari, M.~Sullivan, T.~Tsai, K.~Pattabiraman, J.~Emer, and S.~W.
  Keckler, ``Understanding error propagation in deep learning neural network
  (dnn) accelerators and applications,'' in \emph{SC '17}, 2017, pp. 8:1--8:12.

\bibitem[Pei et~al.(2019)Pei, Cao, Yang, and Jana]{10.1145/3361566}
K.~Pei, Y.~Cao, J.~Yang, and S.~Jana, ``Deepxplore: Automated whitebox testing
  of deep learning systems,'' \emph{Commun. ACM}, vol.~62, no.~11, p.
  137–145, Oct. 2019.

\bibitem[{Deodhare} et~al.(1998){Deodhare}, {Vidyasagar}, and {Sathiya
  Keethi}]{712162}
D.~{Deodhare}, M.~{Vidyasagar}, and S.~{Sathiya Keethi}, ``Synthesis of
  fault-tolerant feedforward neural networks using minimax optimization,''
  \emph{IEEE TNN}, vol.~9, no.~5, pp. 891--900, Sep. 1998.

\bibitem[{Dey} et~al.(2018){Dey}, {Nag}, {Pal}, and {Pal}]{7862272}
P.~{Dey}, K.~{Nag}, T.~{Pal}, and N.~R. {Pal}, ``Regularizing multilayer
  perceptron for robustness,'' \emph{IEEE SMC}, vol.~48, no.~8, pp. 1255--1266,
  Aug 2018.

\bibitem[{Tan} and {Nanya}(1993)]{714236}
Y.~{Tan} and T.~{Nanya}, ``Fault-tolerant back-propagation model and its
  generalization ability,'' in \emph{IJCNN-9}, vol.~3, Oct 1993.

\bibitem[{Sequin} and {Clay}(1990)]{5726611}
C.~H. {Sequin} and R.~D. {Clay}, ``Fault tolerance in artificial neural
  networks,'' in \emph{IJCNN}, June 1990, pp. 703--708.

\bibitem[{Neti} et~al.(1992){Neti}, {Schneider}, and {Young}]{105414}
C.~{Neti}, M.~H. {Schneider}, and E.~D. {Young}, ``Maximally fault tolerant
  neural networks,'' \emph{{IEEE} Trans. Neural Netw.}, vol.~3, no.~1, pp.
  14--23, Jan 1992.

\bibitem[Duddu et~al.(2019)Duddu, Rao, and Balas]{duddu2019adversarial}
V.~Duddu, D.~V. Rao, and V.~E. Balas, ``Adversarial fault tolerant training for
  deep neural networks,'' \emph{arXiv preprint arXiv:1907.03103}, 2019.

\bibitem[bre()]{brevitas}
\BIBentryALTinterwordspacing
 [Online]. Available: \url{https://github.com/Xilinx/brevitas/}
\BIBentrySTDinterwordspacing

\bibitem[Paszke et~al.(2019)Paszke, Gross, Massa, Lerer, Bradbury, Chanan,
  Killeen, Lin, Gimelshein, Antiga, Desmaison, Kopf, Yang, DeVito, Raison,
  Tejani, Chilamkurthy, Steiner, Fang, Bai, and Chintala]{NEURIPS2019_9015}
A.~Paszke, S.~Gross, F.~Massa, A.~Lerer, J.~Bradbury, G.~Chanan, T.~Killeen,
  Z.~Lin, N.~Gimelshein, L.~Antiga, A.~Desmaison, A.~Kopf, E.~Yang, Z.~DeVito,
  M.~Raison, A.~Tejani, S.~Chilamkurthy, B.~Steiner, L.~Fang, J.~Bai, and
  S.~Chintala, ``Pytorch: An imperative style, high-performance deep learning
  library,'' in \emph{NIPS 32}, 2019, pp. 8024--8035.

\bibitem[{Lyons} and {Vanderkulk}(1962)]{5392355}
R.~E. {Lyons} and W.~{Vanderkulk}, ``The use of triple-modular redundancy to
  improve computer reliability,'' \emph{IBM Journal of Research and
  Development}, 1962.

\bibitem[Hinton et~al.(2012)Hinton, Srivastava, Krizhevsky, Sutskever, and
  Salakhutdinov]{hinton2012improving}
G.~E. Hinton, N.~Srivastava, A.~Krizhevsky, I.~Sutskever, and R.~R.
  Salakhutdinov, ``Improving neural networks by preventing co-adaptation of
  feature detectors,'' \emph{arXiv preprint arXiv:1207.0580}, 2012.

\bibitem[zcu()]{zcu104}
\BIBentryALTinterwordspacing
 [Online]. Available:
  \url{https://www.xilinx.com/products/boards-and-kits/zcu104.html}
\BIBentrySTDinterwordspacing

\end{thebibliography}

\end{document}